\documentclass[lettersize,journal]{IEEEtran}
\usepackage{textcomp}

\usepackage{float}
\usepackage{cite}
\usepackage{amsfonts}
\usepackage{textcomp}
\usepackage{mathtools}
\usepackage{amsmath}
\usepackage{amssymb}
\usepackage{graphicx}
\usepackage{booktabs}
\usepackage{multirow}
\usepackage{siunitx}
\usepackage{balance}
\usepackage{colortbl}
\usepackage{lipsum}
\usepackage{hyperref}
\usepackage{mathrsfs}
\usepackage{mathtools}
\usepackage{balance}
\usepackage{multirow}
\usepackage{booktabs}
\usepackage{multirow}
\usepackage{siunitx}
\usepackage{xcolor}
\usepackage{siunitx}
\usepackage{soul}

\usepackage{soul,color}
\usepackage[utf8]{inputenc}
\usepackage{lscape}
\usepackage{rotating}
\usepackage{wrapfig}
\usepackage{url}

\usepackage{lscape} 
\usepackage{caption}
\hypersetup{
	colorlinks=true,
	urlcolor=magenta 
}

\begin{document}
	\title{Deep Generative Adversarial Network for Occlusion Removal from a Single Image}	
		\author{Sanakaraganesh~Jonna, Moushumi~Medhi,
		and~Rajiv~Ranjan~Sahay 
		\thanks{S. Jonna is with the department of Computer Science and Engineering, Indian Institute of Technology, Kharagpur, India, 721302. e-mail: sankar9.iitkgp@gmail.com}
		\thanks{M. Medhi is in the Advanced Technology Development Center, Indian Institute of Technology, Kharagpur, India, 721302. e-mail: medhi.moushumi@iitkgp.ac.in}
		\thanks{R. R. Sahay is with department of Electrical Engineering, Indian Institute of Technology, Kharagpur, India, 721302.}}

	
	
	\maketitle
	
	\begin{abstract}

	Nowadays, the enhanced capabilities of in-expensive imaging devices have led to a tremendous increase in the acquisition and sharing of multimedia content over the Internet. Despite advances in imaging sensor technology, annoying conditions like \textit{occlusions} hamper photography and may deteriorate the performance of applications such as surveillance, detection, and recognition. Occlusion segmentation is difficult because of scale variations, illumination changes, and so on. Similarly, recovering a scene from foreground occlusions also poses significant challenges due to the complexity of accurately estimating the occluded regions and maintaining coherence with the surrounding context. In particular, image de-fencing presents its own set of challenges because of the diverse variations in shape, texture, color, patterns, and the often cluttered environment. This study focuses on the automatic detection and removal of occlusions from a single image. We propose a fully automatic, two-stage convolutional neural network for fence segmentation and occlusion completion. We leverage generative adversarial networks (GANs) to synthesize realistic content, including both structure and texture, in a single shot for inpainting. To assess zero-shot generalization, we evaluated our trained occlusion detection model on our proposed fence-like occlusion segmentation dataset. The dataset can be found on \href{https://github.com/Moushumi9medhi/Occlusion-Removal}{GitHub}.
	
	\end{abstract}

\begin{IEEEkeywords}
	Inpainting, Occlusion Removal, Generative Adversarial Networks
\end{IEEEkeywords}
\vspace{-5pt}

\section{Introduction}

\IEEEPARstart{A}{utomatic}  occlusion removal forms a challenging subpart of the actively pursued occlusion removal problem by the research communities. Image de-fencing in itself, for the most part, has remained an unsolved problem due to variations in shapes, texture, color, patterns, sizes, orientations, and cluttered surrounding vicinities. However, it’s significance cannot be overstated considering its applicability to a myriad of unfavorable imaging conditions, as in capturing images in zoos, museums, gardens, historical landmarks, tourist destinations, or for any other forensic applications, object recognition tasks, etc, where scenes, showpieces, or antiquities are inevitably obstructed by fences or even shadows of frames.
Recently, due to security concerns, some government organizations deliberately corrupt the ID face photos with thin mesh-like structures and named them MeshFace \cite{Zhang_2018_TIFS,Li_2020_PR}. Direct utilization of MeshFace photos results in poor verification performance. In order to improve the performance of recognition systems, they need an inpainting algorithm before verification. De-fencing a single image is strictly an image inpainting problem  wherein a portion of an image to be impainted has to be automatically specified. Image de-fencing involves fence detection and removal of the fence detected in the image by generating occluded background image patches. However, it has often been tackled as a layer or image decomposition-based problem \cite{liu2021learning,chugunov2024neural,zhang2024strong}. \cite{tsogkas2023efficient} has used multiple frames to leverage optical flow between the  frames for de-fencing. Fourier transform was employed in \cite{kume2023single} to detect and remov fences.
In contrast, we present a fully automatic two-stage convolutional neural network for segmentation and completion of fence-like occlusions. In the first stage, we propose a fully convolutional network for occlusion segmentation from a single image. The objective of the second stage is to fill-in missing regions by feeding both input observation and detected occlusion mask. The occlusion segmentation stage is designed by extending UNet \cite{Ronneberger_2015_MICCAI}.

A significant amount of work has been carried out in the literature of image inpainting \cite{Bertalmio_2000_SIGGRAPH,Criminisi_2004_TIP,Guillemot_2014_SPM,Pathak_2016_CVPR,Yu_2018_CVPR,Iizuka_2017_TOG,Liu_2018_ECCV,Yu_2019_ICCV,Zheng_2019_CVPR,Liu_2019_ICCV}. Image completion techniques can be classified into two categories, namely: optimization-based methods \cite{Bertalmio_2000_SIGGRAPH,Criminisi_2004_TIP,Guillemot_2014_SPM} and deep learning-based frameworks \cite{Pathak_2016_CVPR,Yu_2018_CVPR,Iizuka_2017_TOG,Liu_2018_ECCV,Yu_2019_ICCV,Zheng_2019_CVPR,Liu_2019_ICCV}. 
The main challenges in image inpainting are the stable propagation of structures and the simultaneous synthesis of visually meaningful textures to fill in a gap. Inpainting missing regions surrounded by complex backgrounds in images captured in wild or diverse indoor scenarios further adds to the challenges for image inpainting. Traditional methods \cite{Bertalmio_2000_SIGGRAPH, Criminisi_2004_TIP, Papafitsoros_2013_IPOL, Gu_2017_IJCV} for image inpainting usually revolve around some techniques, such as diffusion and exemplar-based, which iteratively search for the best fitting patches to fill in the holes. 
Conventional diffusion-based techniques \cite{Bertalmio_2000_SIGGRAPH,Papafitsoros_2013_IPOL} propagate information from neighboring regions of missing pixels for filling-in occlusions respecting boundaries. Other texture filling methods by Criminisi et al. \cite{Criminisi_2004_TIP} adopted an exemplar-based inpainting technique where available patches are gradually propagated into missing regions. The algorithm in \cite{Gu_2017_IJCV} exploited patch self-similarity through low-rank minimization for image restoration. Due to the unavailability of the high-level semantics, these methods lead to smooth or error-prone  image inpainting results. Also, these techniques fail when the image has a significant amount of missing data. 

In recent years, deep learning has completely accelerated and automated the process of inpainting in real-time. Generative adversarial networks (GANs) \cite{Ian_2014_NIPS} have been extensively used in recent years in image inpainting problems to ensure that the final image obtained after filling in the gaps looks realistic and visually plausible. Some of the notable recent works \cite{Pathak_2016_CVPR,Yu_2018_CVPR,Iizuka_2017_TOG,Liu_2018_ECCV,Yu_2019_ICCV,Zheng_2019_CVPR,Liu_2019_ICCV,shi2022seeing,liu2023coordfill, chugunov2024neural} in image inpainting using deep learning strive to fill the missing holes with reasonable content so that it retains both geometric and inhomogeneous features. Hence, they subdivide the task of image inpainting into multiple tasks such as structure generation and texture generation. In the first stage, a coarsely inpainted output or image structure is produced. In the second stage, a refinement network is employed to synthesize fine textures. However, the two-stage architecture is time-consuming. 

Most of the previous works \cite{Pathak_2016_CVPR,Iizuka_2017_TOG} try to fill a single rectangular hole, often assumed to be the center in the image. Method in \cite{Iizuka_2017_TOG} uses a global GAN that operates on the entire image and additional local GAN on the masked rectangular region to improve results. Models trained solely for rectangular shaped holes may lead to overfitting and limit the applicability of these models in real life. Due to real-time applications, image inpainting with free-form masks \cite{Yu_2019_ICCV} or irregular holes \cite{Liu_2018_ECCV} has drawn attention from many researchers. Hence, in addition to fence-like occlusions, we also consider image occlusions with multiple irregular shapes and sizes, scattered across various locations in the image domain. Unlike previous methods \cite{wu2024syformer}, we aim at recovering both structures and textures in the missing regions of an image at a single shot using a single network that operates robustly on irregular hole patterns to produce semantically meaningful predictions. Using a single generator network  during both the training and testing phases effectively can save inference time.

The completion architecture is composed of three networks – a generator and two discriminator networks.  
 The completion network is trained to fool the discriminator, which requires it to generate images indistinguishable from real ones. The discriminator network is an auxiliary network, used only for training, and can be discarded once the training is completed. During each training iteration, the discriminators are updated first so that they correctly distinguish between real and inpainted training images. Afterwards, the completion network is updated so that it fills the missing area well enough to fool the context discriminator networks.

Since there are not many fence segmentation datasets available for promoting research  in this direction, we   created   the IITKGP\_Fence dataset which includes images with fence-like occlusions, incorporating scale variations, illumination changes, and deformations. The dataset can be accessed at  \url{https://github.com/Moushumi9medhi/Occlusion-Removal}.

\section{Proposed methodology}

In this section, we first present the proposed two-stage algorithm,  consisting of individual networks for fence-like occlusion  segmentation and image inpainting, respectively. The proposed occlusion detection network is built based on UNet \cite{Ronneberger_2015_MICCAI} shown in Fig. \ref{fig:OccNet}. Our occlusion segmentation network takes the whole image as input and outputs the occlusion mask. While the inpainting network takes the input image and the occlusion mask as inputs and produces the inpainted image. 

\subsection{Occlusion segmentation}

\begin{figure*}[!hbt]
	\centering
	\scalebox{1.1}{
		\begin{tabular}{c}
			 
			\includegraphics[width=12cm]{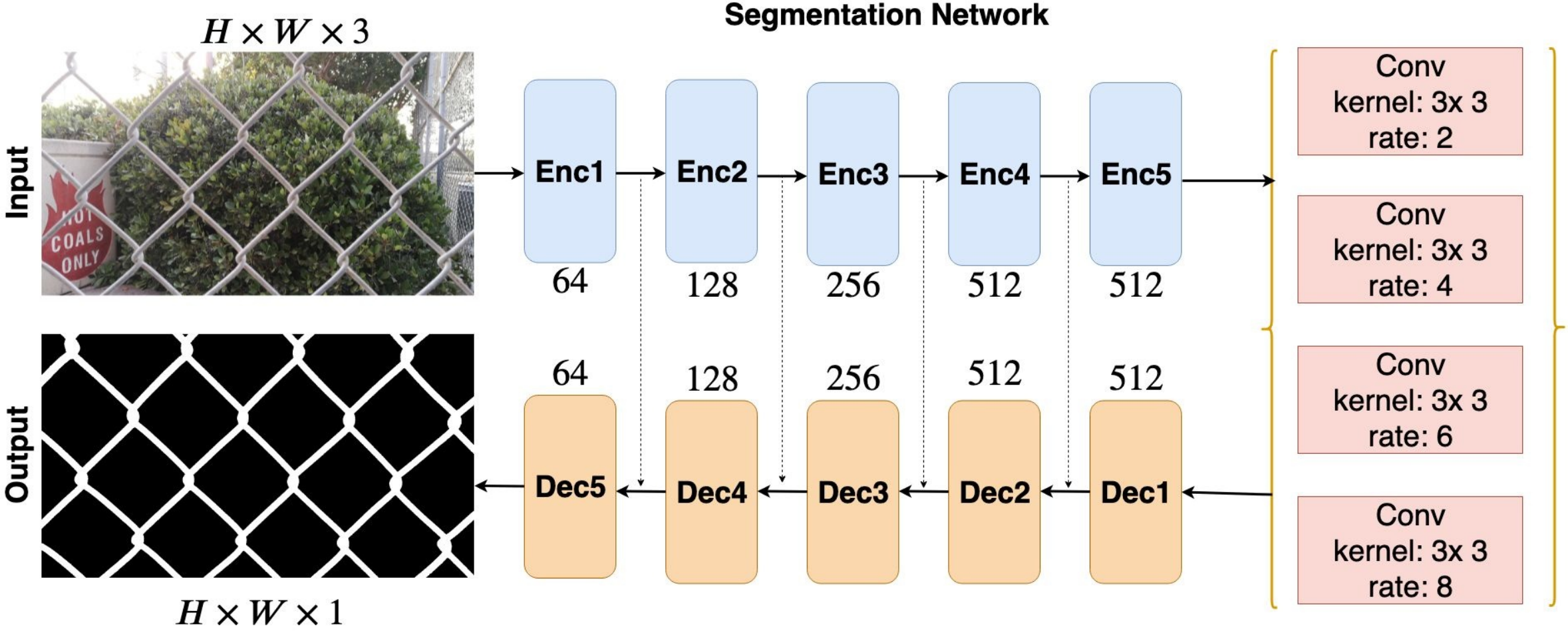}\\
	\end{tabular}}
	\caption{OccNet: Deep fence-like occlusion segmentation architecture.}
	\label{fig:OccNet}	
\end{figure*}

In this study we consider the fence-like occlusion detection as a semantic segmentation problem.
Recent success of fully convolutional networks (FCNs) for various computer vision problems \cite{Shelhamer_2017_PAMI,Badrinarayanan_2017_PAMI,Chen_2018_PAMI,Ronneberger_2015_MICCAI} has motivated us to adopt them for our occlusion segmentation task. The authors in \cite{Shelhamer_2017_PAMI} demonstrated that fully convolutional networks trained end-to-end for semantic segmentation outperform the state-of-the-art. Such end-to-end networks predict dense outputs from arbitrary-sized inputs. Inspired by the successful segmentation networks \cite{Ronneberger_2015_MICCAI,Badrinarayanan_2017_PAMI,Chen_2018_PAMI}, we build a deep occlusion segmentation network, namely, OccNet, based on UNet architecture with an atrous spatial pyramid pooling (ASPP) layer \cite{Chen_2018_PAMI}.   
Atrous convolution allows us to enlarge the field of view of filters to incorporate larger context. 
There are better and more accurate models possible, but in this application of fence-like occlusion detection, we don't need those complexities because of clear boundaries and different textures of background and fence occlusions. Further, this simplicity significantly reduces the time required for the generation of masks by the proposed model. The proposed encoder network is similar to the UNet \cite{Ronneberger_2015_MICCAI} architecture. Each layer in the encoder is a combination of three layers, firstly the convolution layer with batch normalization and ReLU activation. The second one is a convolution layer with input and output of the same dimension, followed by a third $1\times1$ convolution layer with increased output depth. The encoder's output is fed into the atrous convolution layer. The atrous convolution layer returns the sum of results of four atrous convolutions with rate $2, 4, 6$ and $8$. Finally, the decoder upsamples the output of the atrous convolution layer to that of the image input size.

The detailed architecture of the proposed occlusion segmentation network is illustrated in Fig \ref{fig:OccNet}. The encoder shown in Fig. \ref{fig:OccNet} consists of $13$ convolutional layers identical to the VGG16 network in \cite{Simonyan_2015_ICLR}, excluding the last pooling layer and following fully connected layers. This reduces the number of parameters in the encoder part significantly compared to the VGG16 network \cite{Simonyan_2015_ICLR} which was originally designed for object classification. Each module of the encoder network performs convolutions with pre-trained weights to produce feature maps, followed by batch normalization and ReLU layers. It includes max pooling layers with a window size of $2\times 2$ and a stride of $2$ pixels which reduces the size of the input volume to half. The decoder architecture is an exact mirror image of the encoder with max pooling layers being replaced by upsampling layers (transposed convolution). Similar to the UNet \cite{Ronneberger_2015_MICCAI}, the decoder blocks get features from the corresponding encoder blocks via skip-connections. The encoder and decoder features are concatenated together. The decoder part uses upsampling layers followed by convolution layers to obtain the output, which is at the same resolution as the input. Here, upsampling is performed in-network for end-to-end learning by backpropagation. Similar to the encoder part, ReLU non-linearities are used in all the convolutional layers with a final sigmoidal non-linearity layer to produce the segmentation map. 

Given a training set of input-target pairs ${\mathbf{X}^{(i)},\mathbf{Y}^{(i)}}$ for the supervised occlusion  segmentation task, our objective is to learn the parameters $\theta$ of a representation function $G_\theta$ which optimally approximates the input-target dependency according to a loss function $\mathscr{L}(G_{\theta}(\mathbf{X}),\mathbf{Y})$. In the proposed encoder-decoder network, we use the binary cross entropy loss function, which is the average of the individual binary cross entropies (BCE) across all $N$ pixels.
\begin{equation}
\mathscr{L}_{BCE} = -\frac{1}{N} \sum_{j=1}^{N}\mathbf{Y}_j\log(\hat{\mathbf{Y}}_j) + (1-\mathbf{Y}_j)\log(1-\hat{\mathbf{Y}}_j)
\end{equation}

To train this CNN, we follow the dataset splitting scheme of \cite{Du_2018_ICME}. Before training the network, decoder weights are initialized randomly. The encoder part of the network is initialized with the weights of a VGG16 model \cite{Simonyan_2015_ICLR} trained on the ImageNet \cite{Deng_2009_CVPR} dataset for object recognition. The learning rate is changed over epochs, starting with a value of $1\times 10{-3}$. We trained the network for $100$ epochs. During the testing phase, we feed the input image through the trained occlusion segmentation model  and obtain the occlusion mask of the same resolution as that of the input.

\subsection{Image inpainting}

Generative Adversarial Networks (GANs) \cite{Ian_2014_NIPS} are  frameworks for training generative parametric models, and have been shown to produce high quality images. This framework trains two networks simultaneously, a generator, $G$, and a discriminator, $D$. $G$ maps a random vector $z$, sampled from a prior distribution, $p_z$, to the image space while $D$ maps an input image to a likelihood. The purpose of $G$ is to generate realistic images, while $D$ plays an adversarial role, discriminating between the image generated from $G$, and the real image sampled from the data distribution $ p_{data} $. With some user interaction, GANs have been applied in interactive image editing. However, GANs can not be directly applied to the inpainting task, because they produce an entirely unrelated image with high probability, unless constrained by the provided corrupted image.
\linebreak[2]

\begin{figure*}[!hbt]
	\centering
	\scalebox{1.1}{
		\begin{tabular}{c}
			\includegraphics[width=12.5cm]{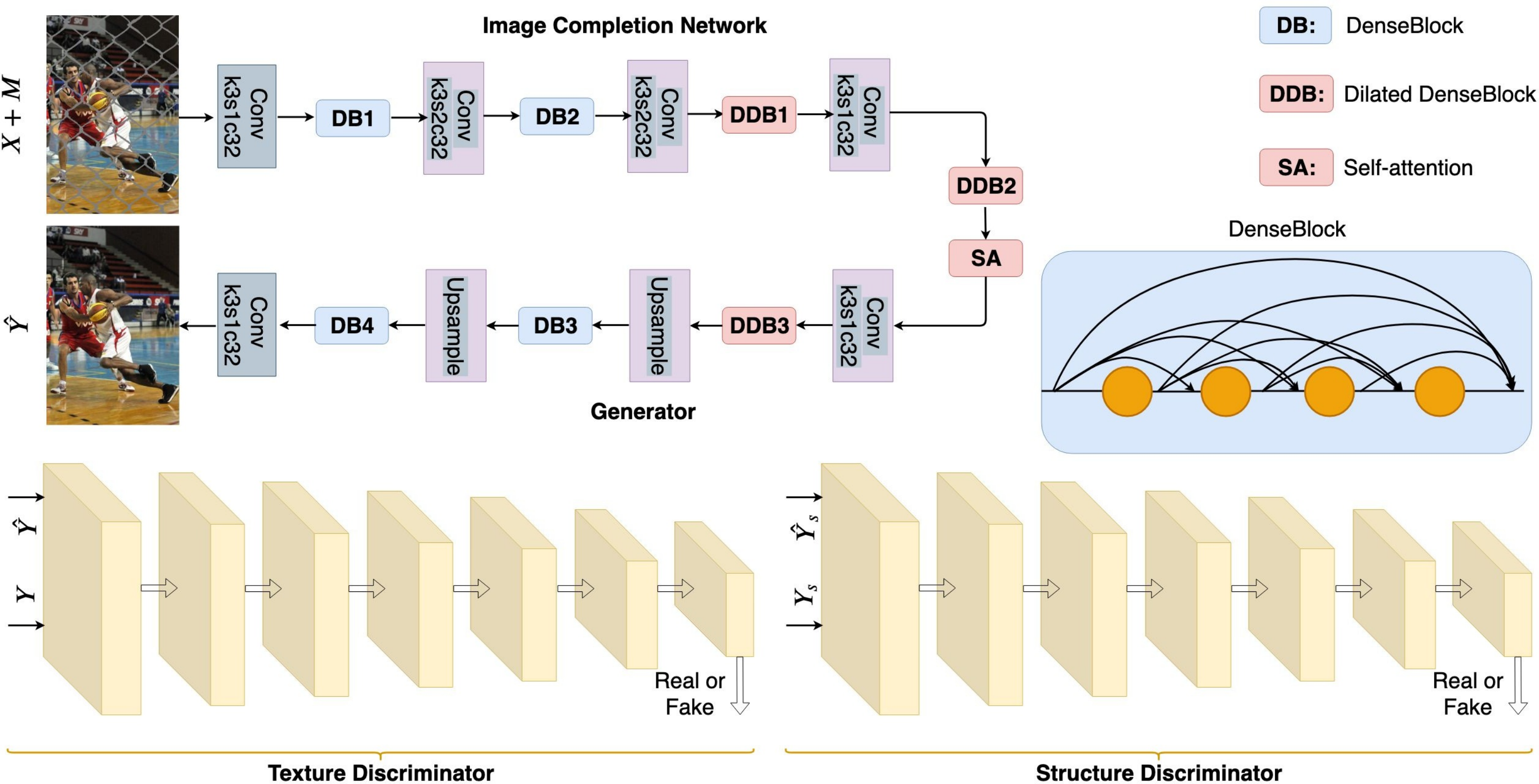}\\
	\end{tabular}}
	\caption{Schematic of the proposed image inpainting network.}
	\label{fig:CoarseNet}	
\end{figure*}

There are several variants of GANs proposed in the literature, such as vanilla GAN with patch discriminator \cite{Isola_2017_CVPR}, Wasserstein GANs (WGAN) \cite{Gulrajani_2017_NIPS}, and least squares generative adversarial networks (LSGAN) \cite{Mao_2019_PAMI}. Regular GANs suffer from vanishing gradient problem during training  due to sigmoid cross-entropy loss function. In this work, we adopted LSGAN \cite{Mao_2019_PAMI} due to its stability and high performance during training.
The complete schematic of the proposed image completion framework is shown in Fig. \ref{fig:CoarseNet}. 
Our GAN framework consists of a generator network (inpainting network) and two auxiliary discriminator networks: texture discriminator and structure discriminator, that are used only during training and not during testing. Generator network takes an input image $X$ along with the occlusion binary mask. 
The texture discriminator forces the generator to produce visually realistic contents in the missing regions, whereas the structure discriminator induces the generator to plausibly reconstruct the structure.

\begin{figure*}[!hbt]
	\centering
	\scalebox{0.85}{
		\begin{tabular}{c c c c}
			\includegraphics[width=3.2cm]{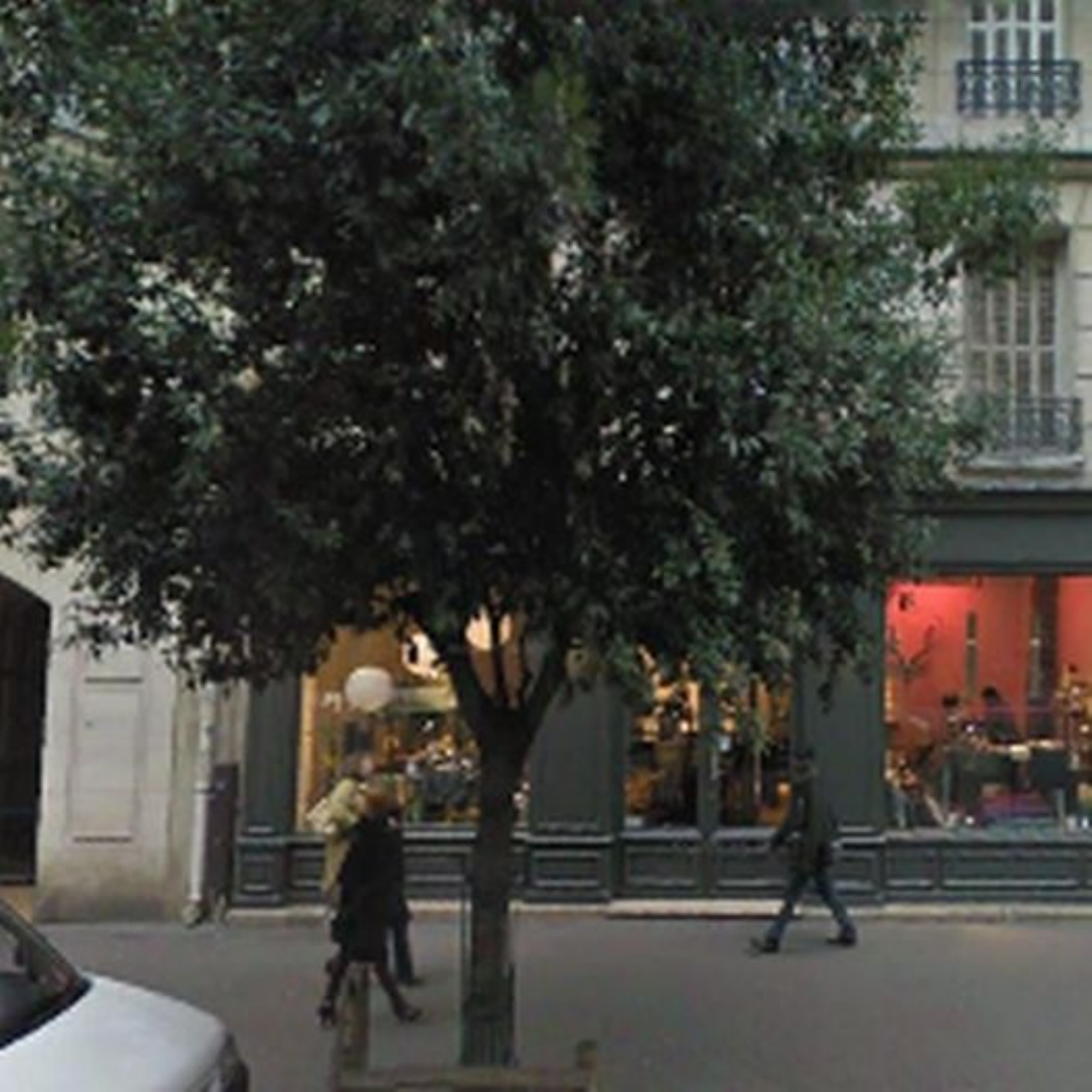}&\hspace{-12pt}
			\includegraphics[width=3.2cm]{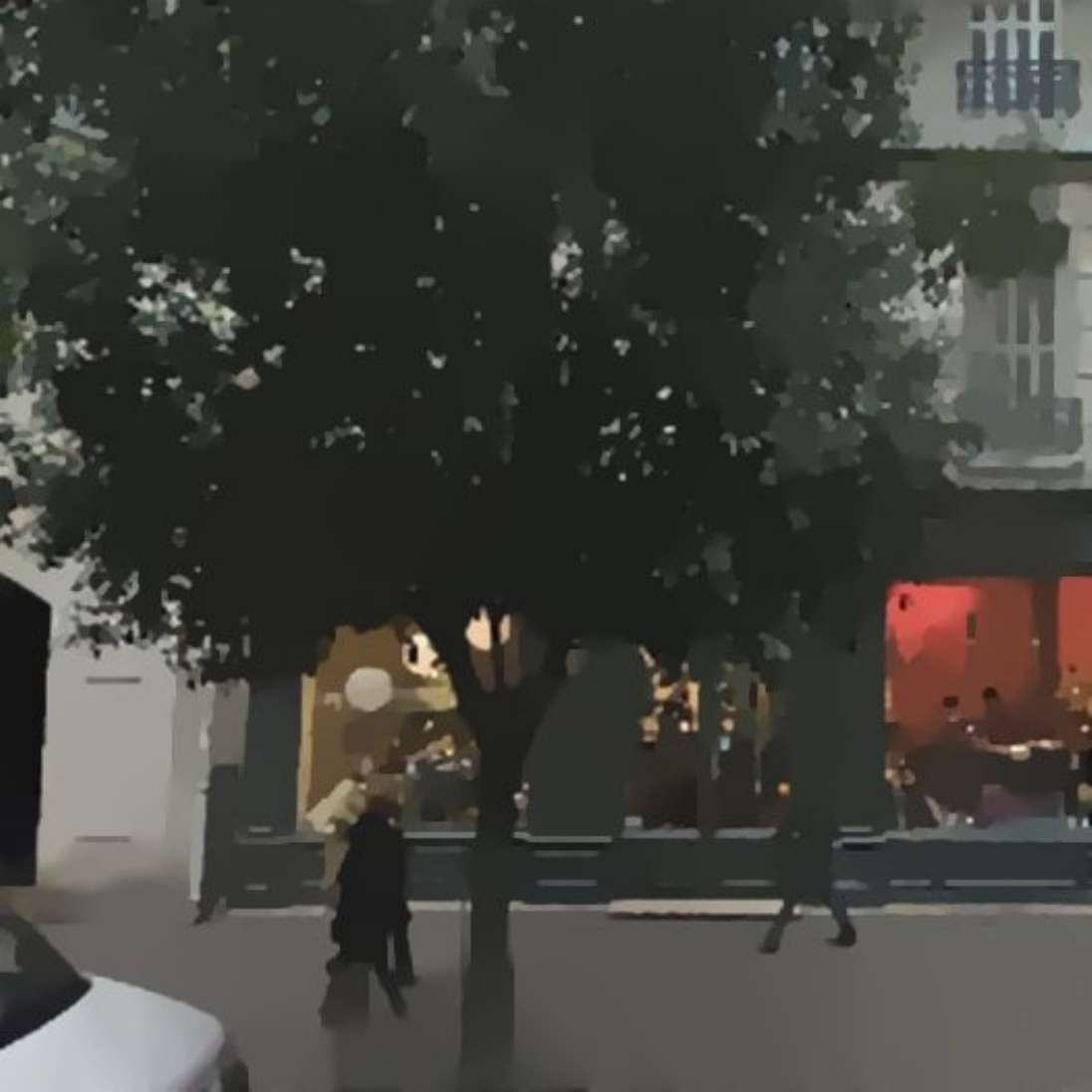}&\hspace{-12pt}
			\includegraphics[width=3.2cm]{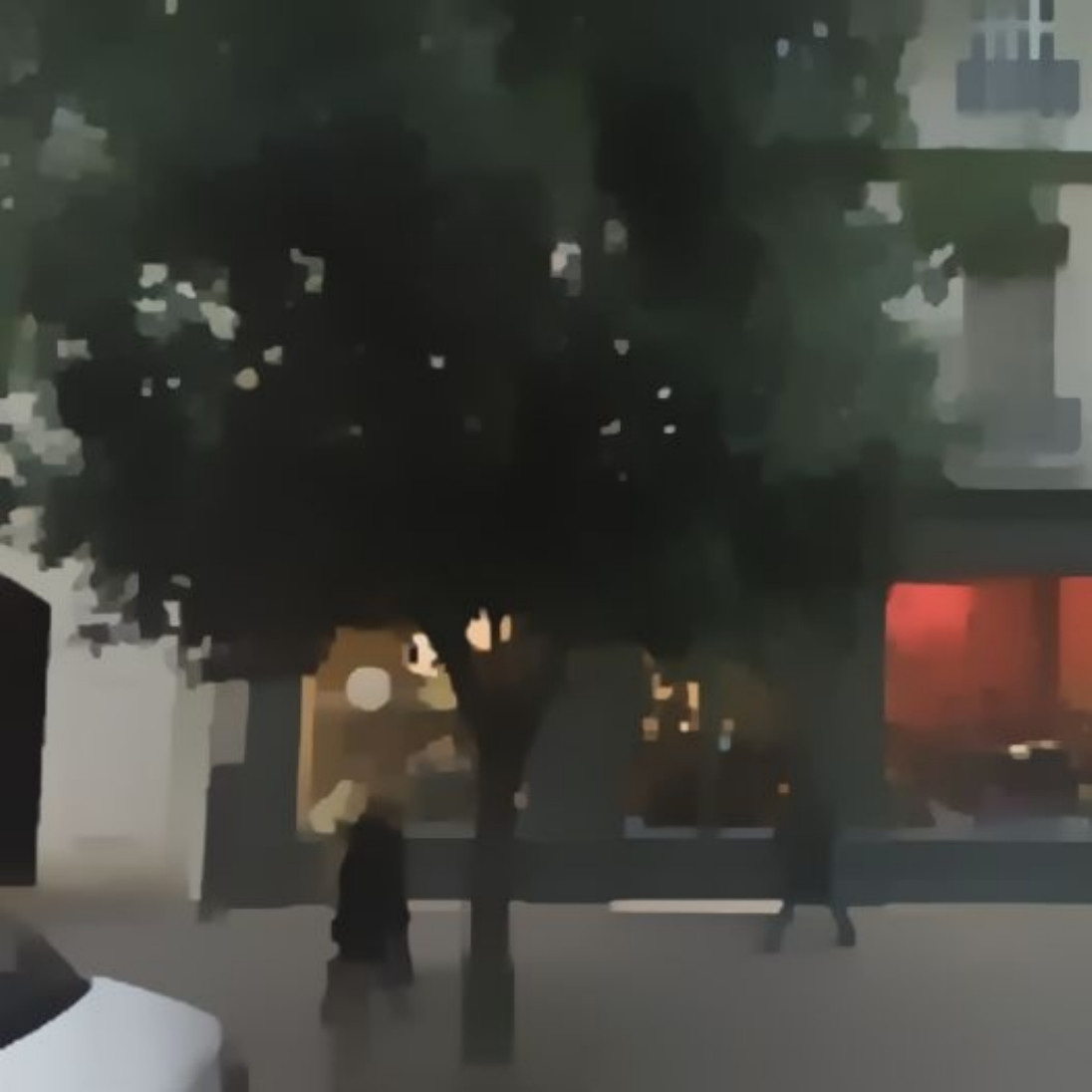}&\hspace{-12pt}
			\includegraphics[width=3.2cm]{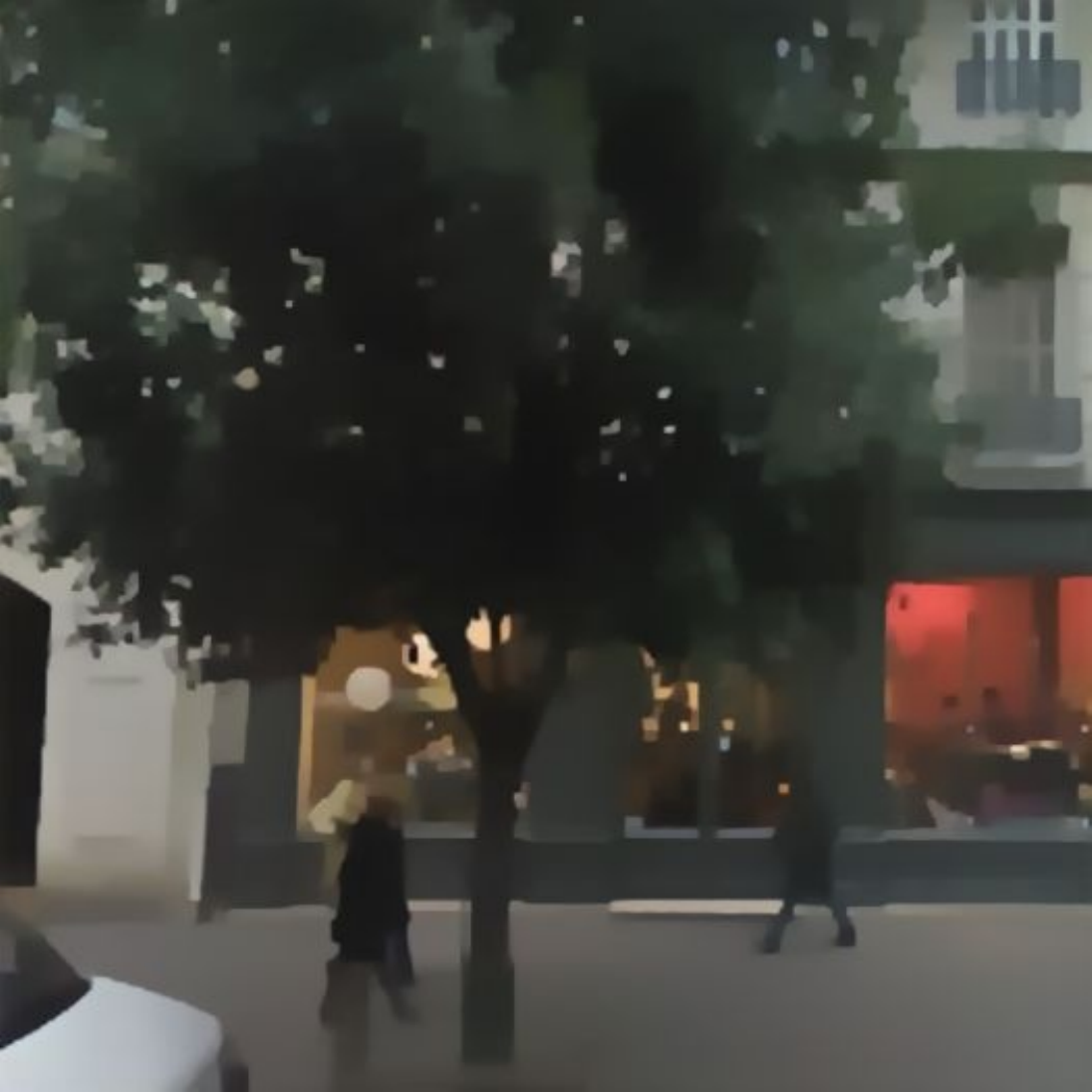}\\
			(a) & (b) & (c) & (d)\\	
		\end{tabular}}
		\caption{(a) Input image. (b)-(d) Edge-preserving structures obtained using $L0$ \protect \cite{Xu_2011_TOG}, RTV \protect \cite{Xu_2012_TOG}, and the pretrained model in \protect \cite{Fan_2019_PAMI}, respectively.}
		\label{fig:smooth_variants}	
	\end{figure*}

The structures of an image can be obtained using various traditional optimization-based methods such as, non-local means filtering (NLM) \cite{Buades_2005_CVPR}, $L0$ minimization \cite{Xu_2011_TOG}, relative total variation (RTV) \cite{Xu_2012_TOG}. However, such methods are time consuming and cannot be applied for real time scenarios. We therefore procure the structures of the prediction and the ground truth on the fly from a decouple learning-based pretrained model \cite{Fan_2019_PAMI} which was trained on an RTV image operator. The image structures obtained using various filtering operators are shown in Fig. \ref{fig:smooth_variants}.
Also, note that we do not employ any additional post-processing, blending operation, or refinement networks. The input image with missing regions and the corresponding mask is fed to the network, which then undergoes several convolutions and deconvolutions as it traverses across the network layers.

\subsubsection{Generator architecture}

Our generator network $G$  is built on an encoder-decoder like structure. Dense convolutional blocks \cite{Huang_2017_CVPR} have proved their ability to solve the  gradient vanishing problem, which commonly occurs in deep networks. Our encoder includes several blocks containing dense connections. The initial layers of the encoder have two dense blocks (DB), each followed by a max-pooling layer. Since a larger field of view is required for the image completion problem, the dense blocks in the deeper layers are equipped with dilated convolutional layers dense blocks (DDB),  referred to as dilated dense blocks (DDB). We use  dilation rates of  $2$, $4$, $6$, and $8$. In order to include global contextual information in addition to local pixels, we include the self-attention (SA) module \cite{Ashish_2017_NIPS,Wang_2018_CVPR,Zhang_2019_ICML} in between the encoder and decoder parts. Each layer in the proposed generator model $G$ uses gated convolutions \cite{Yu_2019_ICCV} to learn contributing features at each spatial location.

Architectures that use vanilla convolutional layers naturally treat all input pixels as valid ones. However, in image inpainting problem, the input images/features consist of both regions with visible pixels outside the mask and missing or generated pixels in the missing regions. This equal treatment by standard vanilla convolutional layers leads to structural or textural artifacts and blurriness in  the reconstructed regions. In order to address the issue, the authors in \cite{Liu_2018_ECCV} proposed a partial convolutional layer (PConv), which is designed to handle missing or occluded regions in images by applying convolutions only to valid pixels and dynamically adjusting the convolutional kernels based on the presence of valid data, thereby improving the quality of image inpainting. Subsequently, the authors in \cite{Yu_2019_ICCV} proposed a generalized convolutional layer called gated convolutional layer (GConv), which learns soft masks automatically from the data, rather than relying on hard masks like those used in PConv:
\begin{equation}
\begin{split}
Gating_{x,y} = \sum\sum W_g\cdot F_{x,y} \\
Features_{x,y} = \sum\sum W_f\cdot F_{x,y} \\
Output_{x,y} = \phi(Features_{x,y})\odot\sigma(Gating_{x,y})
\end{split}
\end{equation}
where $F_{x,y}$ denotes input features, $\phi$ can be any activation functions, $\sigma$ is the sigmoid function that generates output gating values between zeros and ones. $W_g$ and $W_f$ are two different convolutional filters.

Each layer of the generator network $G$ uses gated convolutions \cite{Yu_2019_ICCV} to learn contributing features at each spatial location. The generator network takes an image with gray colored missing pixels to be inpainted and a binary mask indicating the missing regions as input pairs and outputs the final inpainted image. We pair the input with a corresponding binary mask to handle holes with variable sizes, shapes, and locations. 

\subsubsection{Self-attention module}

\begin{figure}[!hbt]
	\centering
	\scalebox{1.1}{
		\begin{tabular}{c}
			\includegraphics[width=7.5cm]{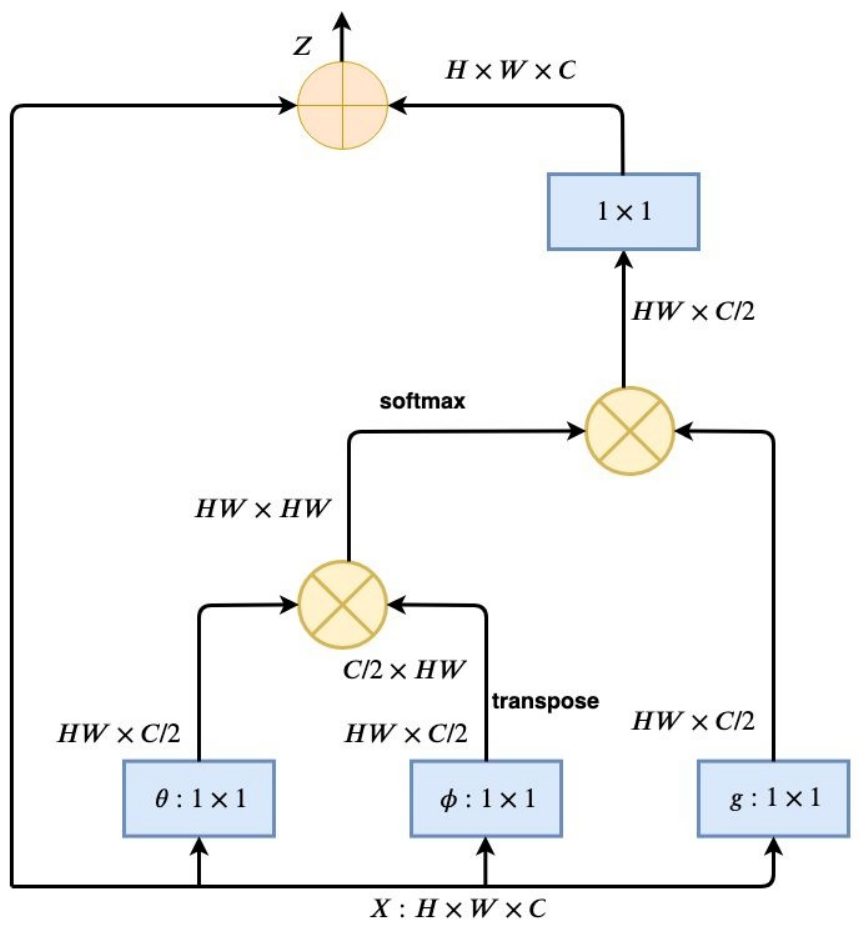}\\
	\end{tabular}}
	\caption{Self-attention module \protect \cite{Ashish_2017_NIPS,Wang_2018_CVPR}.}
	\label{fig:CA}	
\end{figure}

Long-range dependency is effective in enhancing inpainting tasks and can be achieved by considering large receptive fields, which is possible by stacking multiple convolutional layers. However, stacking layers may suffer from two major limitations: 1) computationally expensive, 2) difficulties in training. Inspired by the classical non-local means algorithm in \cite{Buades_2005_CVPR}, the authors in  \cite{Wang_2018_CVPR} proposed a generic non-local module for modeling long-range dependencies. In non-local operation \cite{Buades_2005_CVPR}, each pixel is replaced by the weighted average of the similar pixels taken from the whole image.
Inspired by \cite{Buades_2005_CVPR,Ashish_2017_NIPS}, and \cite{Wang_2018_CVPR}, we built the proposed generator architecture with a self-attention block. Normal convolutional layers process the data in a local neighborhood. The self-attention layer learns where to borrow or copy feature information from known background patches to generate missing patches. It is differentiable, allowing it to be trained in deep models, and fully convolutional, which enables testing at arbitrary resolutions. The self-attention block is shown in Fig. \ref{fig:CA}.

\subsubsection{Structure and texture discriminators}
For the simultaneous recovery of textures and structures, our model consists of two discriminators, namely: texture discriminator $D_t$ and structure discriminator $D_s$. Both the discriminators were   designed with the same architectural components. Each consists of $7$ convolutional layers and a single fully connected layer that outputs a single-dimensional vector. All the convolutional layers employ a striding of $2\times 2$ pixels to decrease the feature resolution. In contrast with the inpainting network, all convolutions use $4\times 4$ kernels. For the stable training of GANs, we employ a recent weight normalization strategy called spectral normalization (SN) \cite{Miyato_2018_ICLR}. Although there are many variants of GANs, such as Vanilla GAN \cite{Ian_2014_NIPS}, Vanilla GAN with patch discriminator \cite{Isola_2017_CVPR}, Wasserstein GANs (WGAN), etc, we adopted the least squares GAN (LSGAN) \cite{Mao_2019_PAMI}  for stable and faster training while minimizing the least squares losses. The input for the texture discriminator is either ground truth image or prediction from the generator $G$. Structural discriminator takes structure obtained from pre-trained model \cite{Fan_2019_PAMI} by feeding prediction from $G$ or the ground truth image. The detailed architecture of the texture discriminator or structure discriminator is shown in Table \ref{tab:Disc}. 


\begin{table}[!htb]
	\centering
	\caption{Architecture of the discriminator network.}
	\scalebox{1.1}
	{
		\begin{tabular}{c c c c c c c}
			\hline
			
			Type & Kernel & Dilation & Stride & Padding & Outputs\\
			\hline
			\hline
			SNConv & $4\times 4$ & 1 & $2$ & $2$ & 64\\
			SNConv & $4\times 4$ & 1 & $2$ & $2$ & 128\\
			SNConv & $4\times 4$ & 1 & $2$ & $2$ & 256\\
			SNConv & $4\times 4$ & 1 & $2$ & $2$ & 256\\
			SNConv & $4\times 4$ & 1 & $2$ & $2$ & 256\\
			SNConv & $4\times 4$ & 1 & $2$ & $2$ & 256\\
			SNConv & $4\times 4$ & 1 & $2$ & $2$ & 256\\									
			\hline						
	\end{tabular}}
	\label{tab:Disc}
\end{table}

\subsection{Loss function}
\noindent
\textbf{Reconstruction loss:}  We compute pixel-wise $\mathscr{L}_1$ loss between the prediction and the ground truth image. For a given occluded image $X$ along with the occlusion mask $M$, we obtain the  inpainted image $\hat{Y}$ from our generator $G$. The pixel-wise $\mathscr{L}_1$ loss between $\hat{Y}$ and ground truth $Y$ is defined as
\begin{equation}
	\mathscr{L}_{rec} = \parallel \hat{Y} -Y\parallel_1
\end{equation}
\noindent
\textbf{Perceptual loss:}  Models trained solely with  pixel-wise reconstruction loss often lead to blurry predictions in masked regions. To enhance the details of predictions,  several inpainting and style transfer networks incorporate feature loss. Following this approach, we compute perceptual and style losses using a pre-trained VGG model. We consider the first three pooling layers for the perceptual loss as follows:
\begin{equation}
	\mathscr{L}_{per} = \frac{1}{N}\sum_{i=1}^{N}\parallel \theta_i(\hat{Y}) -\theta_i(Y)\parallel_1
\end{equation}
\noindent
\textbf{Structure loss:}  Unlike many dual generator-based models \cite{Nazeri_2019_ICCVW,Ren_2019_ICCV,Xiong_2019_CVPR} that separately address structure and texture recovery, our framework computes the structure loss directly between the prediction and the ground truth.  Given the high computational cost of traditional non-learning-based image operators \cite{Xu_2012_TOG}, we consider the recent pre-trained model \cite{Fan_2019_PAMI} in place of the edge-preserving smoothness operator. Let $G_s$ denotes the structure generation model, $\hat{Y}^s$ and  $Y^s$ denote the predicted and the ground truth structures obtained from $G_s$.The structure loss is computed as follows: 
\begin{equation}
\mathscr{L}_{str} = \parallel G_s(\hat{Y}) - G_s(Y)\parallel_1
\end{equation}
\noindent
\textbf{Texture discriminator loss:}
\begin{equation}
\mathscr{L}_{D_t} = \frac{1}{2}E((D_t(Y)-1)^2 + D_t((G(X,M))^2))
\end{equation}
\noindent
\textbf{Structure discriminator loss:}
\begin{equation}
\mathscr{L}_{D_s} = \frac{1}{2}E((D_s(G_s(Y))-1)^2 + D_t((G_s(G(X,M)))^2))
\end{equation}
The overall objective function is given as follows:
\begin{equation}
\mathscr{L} = \lambda_1 \mathscr{L}_{rec} + \lambda_2 \mathscr{L}_{per} + \lambda_3 \mathscr{L}_{str} + \lambda_4 \mathscr{L}_{D_t}  + \lambda_5 \mathscr{L}_{D_s}
\label{eq:loss}
\end{equation}

\section{Experiments}
\label{sec:exp}

\subsection{Occlusion segmentation}
\subsubsection{Datasets}
\begin{figure}[!hbt]
	\centering
	\scalebox{0.605}{
		\begin{tabular}{c c c c}
			\includegraphics[width=3.3cm]{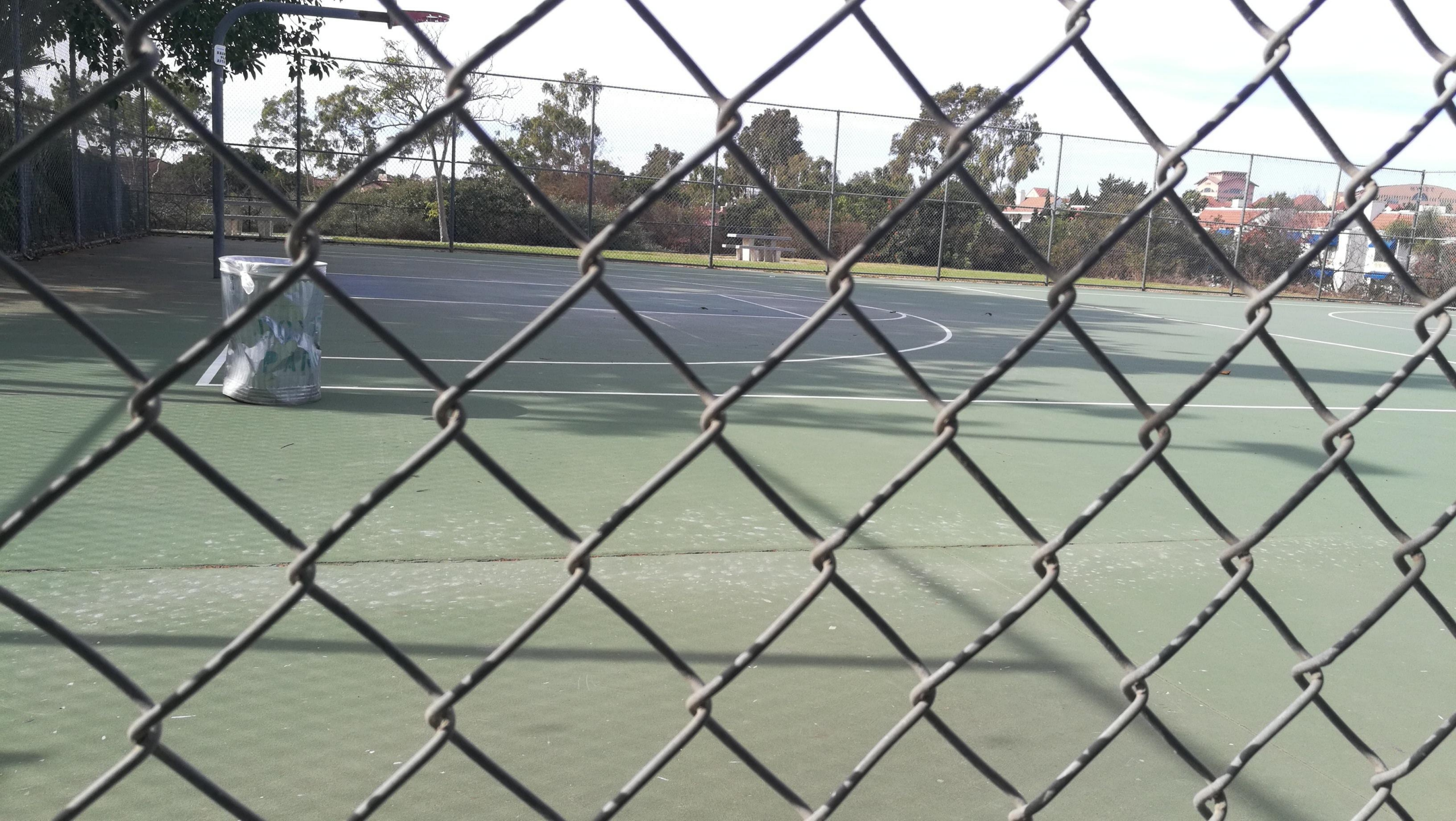}&\hspace{-12pt}
			\includegraphics[width=3.3cm]{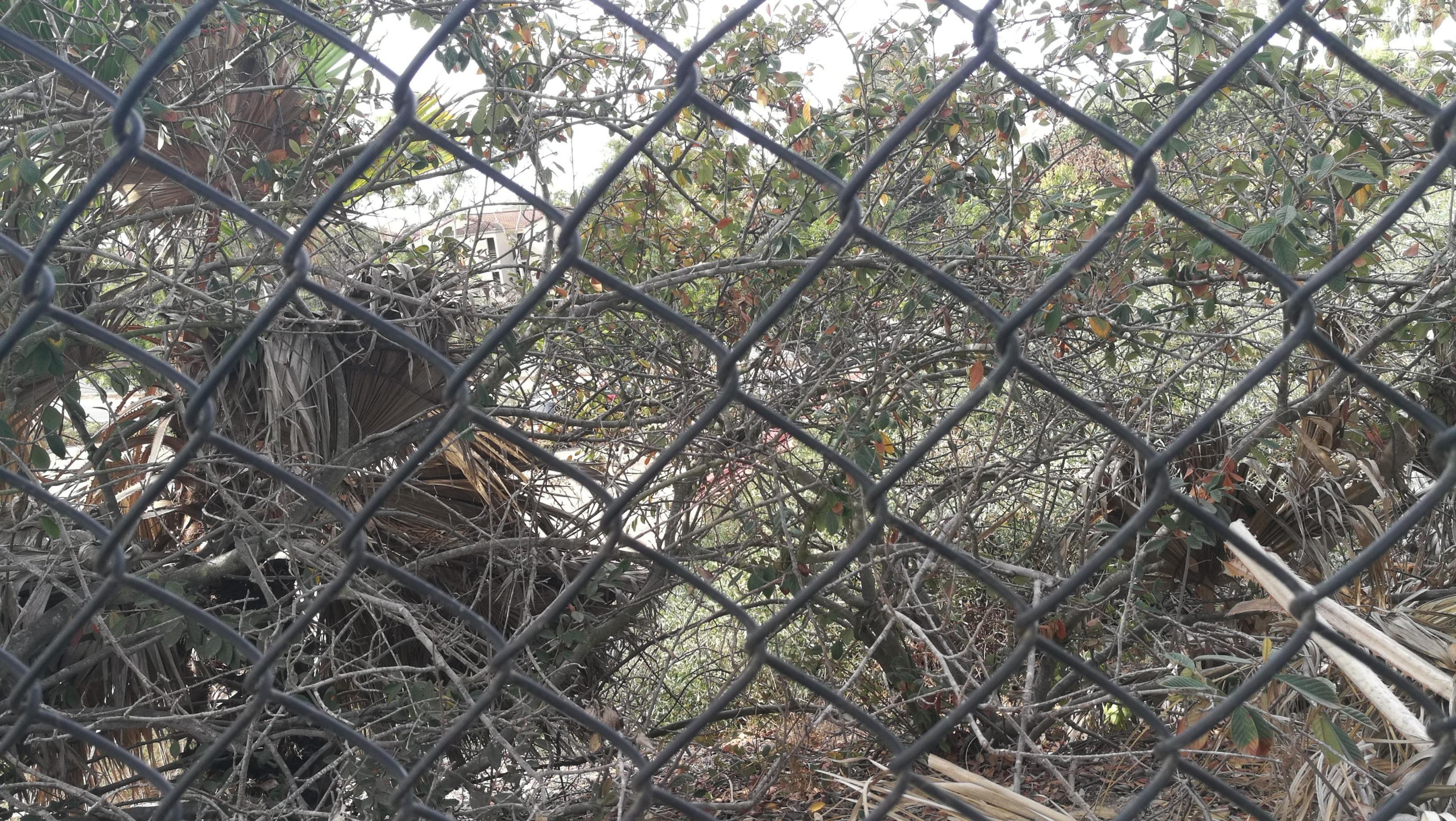}&\hspace{-12pt}	
			\includegraphics[width=3.3cm]{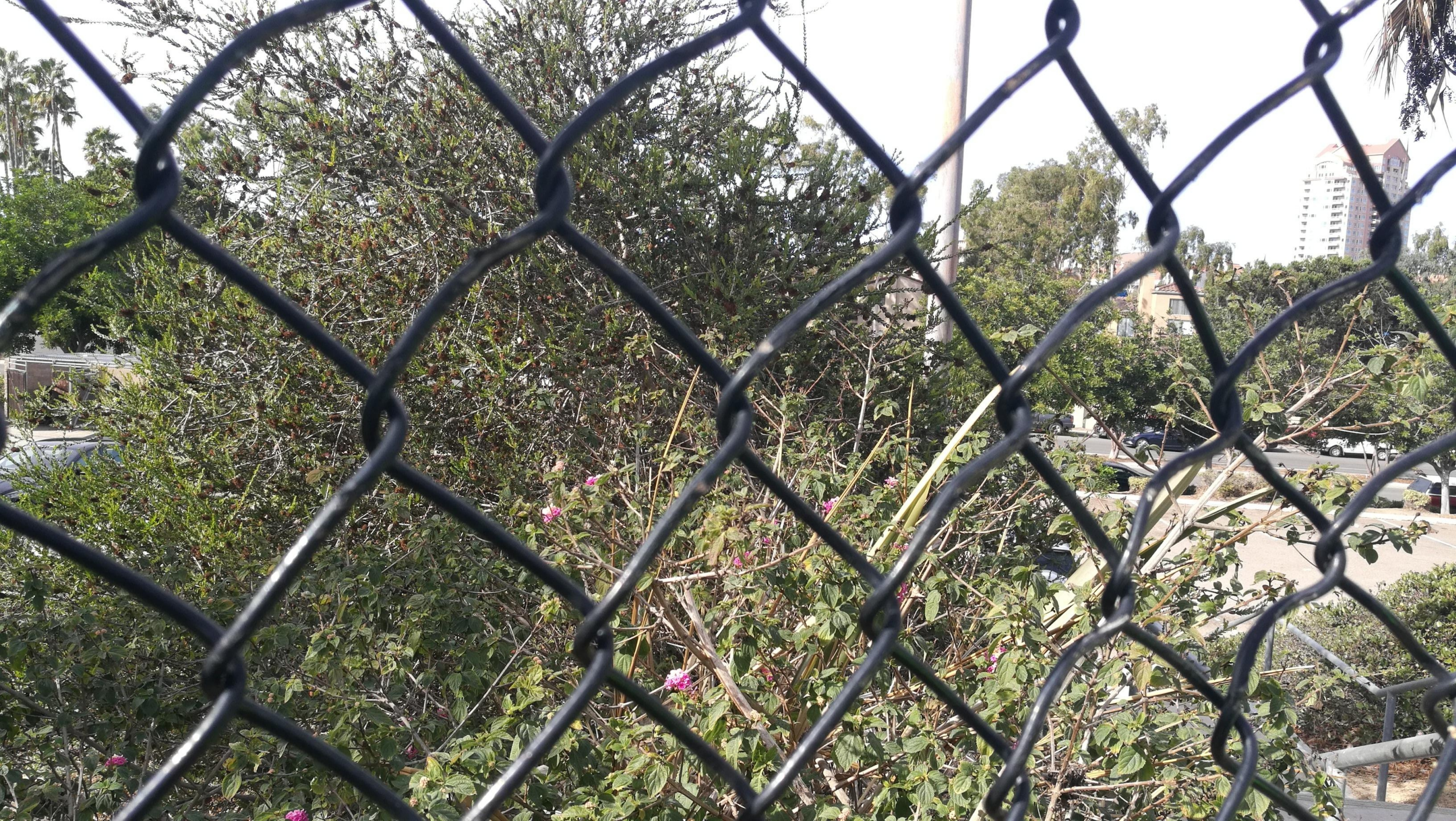}&\hspace{-12pt}
			\includegraphics[width=3.3cm]{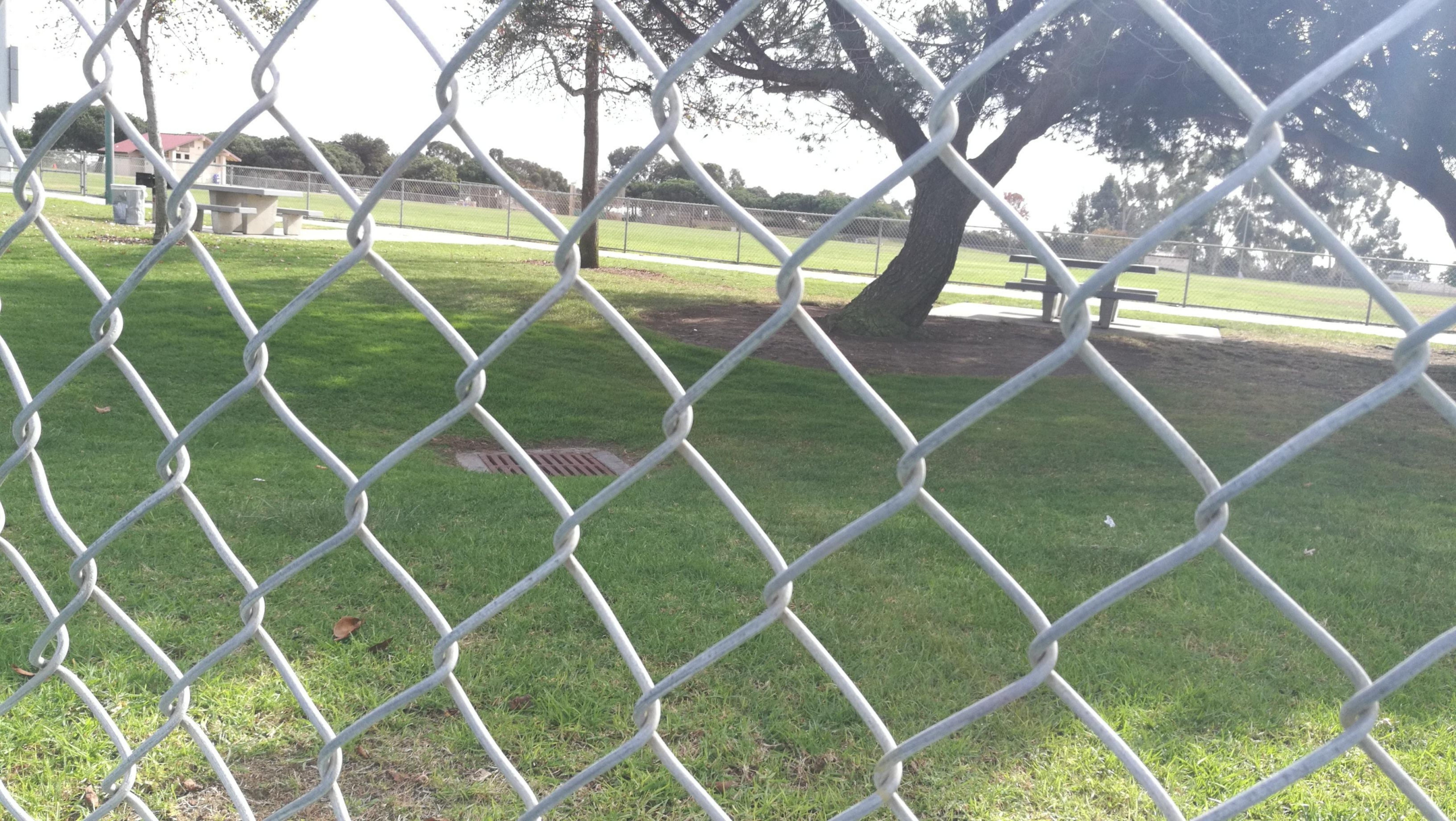}\\
	\end{tabular}}
	\caption{Sample fence images from UCSD\_Fence dataset \protect \cite{Du_2018_ICME}.}	
	\label{fig:sample_ucsd}
\end{figure}

We consider two datasets for evaluating the proposed fence-like occlusion segmentation network, namely: UCSD$\_$Fence dataset and the IITKGP$\_$Fence dataset.

\noindent \\
\textbf{UCSD\_Fence dataset \cite{Du_2018_ICME}:}
The UCSD\_Fence dataset consists of $645$ pairs of  images and the corresponding ground truth fence masks. The train and the test splits consist of $545$ and $100$ images, respectively. We follow the splitting scheme in \cite{Du_2018_ICME} to train the proposed framework. The images were resized to $320\times 320$ before feeding to the network. Some samples of the fence occluded images from the UCSD\_Fence dataset are shown in Fig. \ref{fig:sample_ucsd}.

\begin{figure}[!hbt]
	\centering
	\scalebox{1.1}{
		\begin{tabular}{c c c}
			\includegraphics[width=2.5cm]{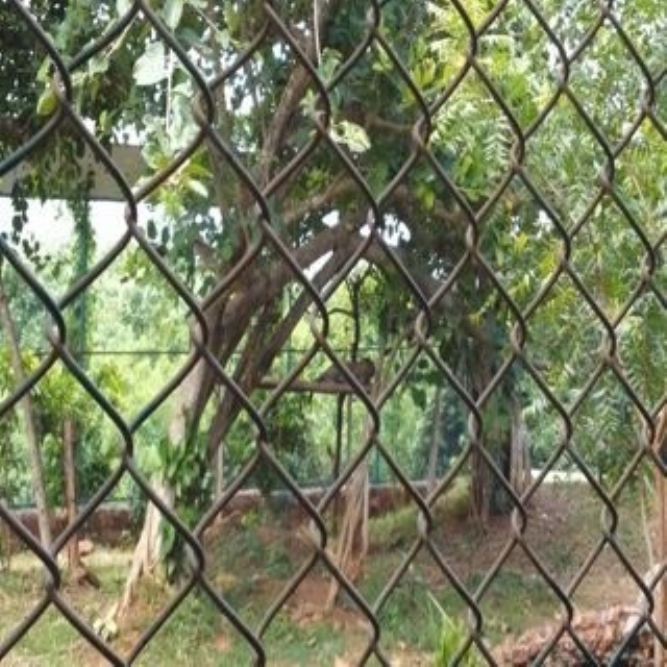}&\hspace{-12pt}
			\includegraphics[width=2.5cm]{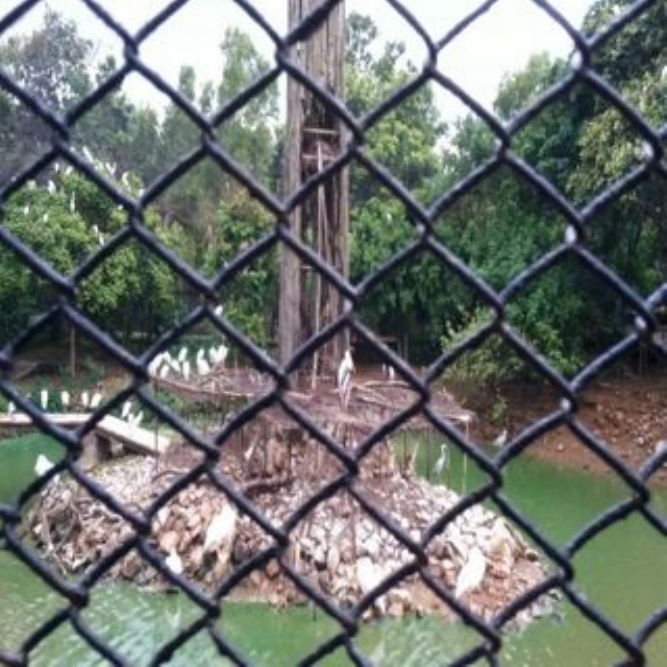}&\hspace{-12pt}
			\includegraphics[width=2.5cm]{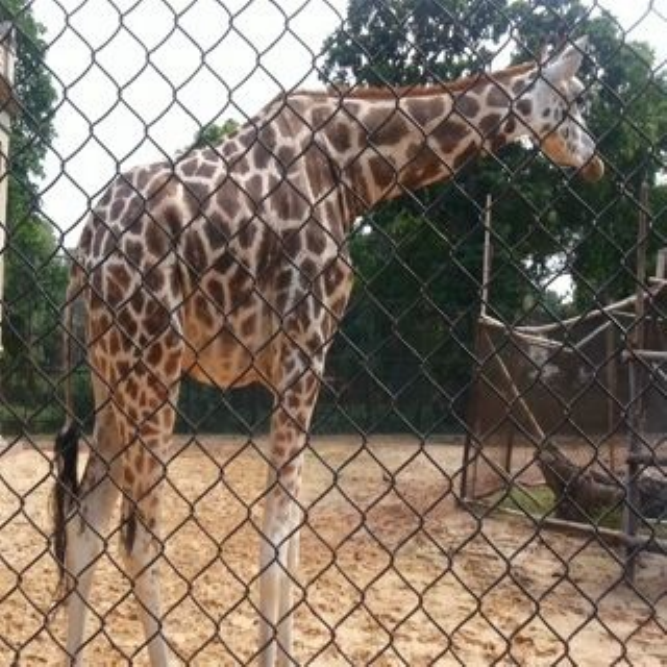}\\
	\end{tabular}}
	\caption{Sample fence images from IITKGP\_Fence dataset.}
	\label{fig:sample_iitkgp}	
\end{figure}

\noindent \\
\textbf{IITKGP\_Fence dataset:} We use our open dataset, IITKGP\_Fence, which includes variations in scene composition, background defocus, and object occlusions, offering diverse conditions for fence-like occlusion detection and evaluation. The dataset comprises both labeled and unlabeled data, as well as additional video and RGB-D data. In this work, we utilized a subset of 175 labeled images. We created the ground truth labels for the proposed dataset  in a semi-automatic way with user interaction. In this study, we used this dataset only for evaluation purposes and not for training. For training the OccNet, we rather chose to train using an established publicly available dataset \cite{Du_2018_ICME} for fair comparison with existing methods.  We created the dataset to address the deficiency of fence-like occlusion datasets and establish a new benchmark for assessing the performance of existing models. Evaluating a model on a dataset not utilized during training ensures an unbiased assessment and offers a true measure of its performance on novel data. In Fig. \ref{fig:sample_iitkgp}, we show a few sample images from IITKGP\_Fence dataset.

\subsubsection{Evaluation metrics}
We use the standard quantitative metrics such as precision-recall (PR) curves, F-measure, and mean absolute error (MAE) to evaluate our proposed deep OccNet. The precision and the recall values are obtained by thresholding the predicted fence map between $0$ to $255$, and comparing it with the ground truth. Precision and recall values are calculated as follows: Precision = $\frac{|B\cap G|}{|B|}$,  Recall = $\frac{|B\cap G|}{|G|}$, where $|$$\cdot$$|$ gathers the non-zero entries in a mask, and $B$, $G$, are binary-valued predicted and ground-truth fence maps, respectively. PR curves demonstrate the mean precision and recall values over a dataset at various thresholds. To assess the quality of the predicted fence segment, a combined F-measure metric is used, which is defined as:
\begin{equation}
F\text{-}measure= \frac{(1+\beta^2) \times precision \times recall}{\beta^2\times precision + recall}
\end{equation}
As recommended by several previous works, we set the value of $\beta^2$  to  $0.3$ to give more importance to precision. We present the F-measure curves computed at different thresholds for each dataset. For performance evaluation, we binarize each fence map using a data-dependent adaptive threshold, determined as the mean value of the fence map:
$T_{a} = \frac{k}{W\times H}\sum\limits_{i=1}^{W}\sum\limits_{j=1}^{H}B(i,j)$
where $k=1.0$; $W$ and $H$ represent the width and height of the fence map $B$, and $B(i, j)$ is the fence value of the pixel at $(i, j)$. Given the fence map $B$ and the ground truth mask $G$, the mean absolute error (MAE) can be calculated as,
\begin{equation}
MAE = \frac{1}{W\times H} \sum\limits_{i=1}^{W}\sum\limits_{j=1}^{H}|B(i,j)-G(i,j)|
\end{equation}

\subsubsection{Quantitative and Visual Results}

We report both the quantitative and visual results obtained from the existing and   proposed methods. The quantitative results  from  the existing and proposed methods on our IITKGP\_Fence dataset are shown in Table \ref{tab:quant_fence1}, while the results  on the UCSD\_Fence dataset are presented in Table \ref{tab:quant_fence2}. Note that the proposed OccNet outperforms other algorithms, as demonstrated in the last rows of Tables \ref{tab:quant_fence1} and \ref{tab:quant_fence2}.

\begin{figure*}[!hbt]
	\centering
	\scalebox{0.8}{
		\begin{tabular}{c c c c c c c c}
			\includegraphics[width=2.0cm]{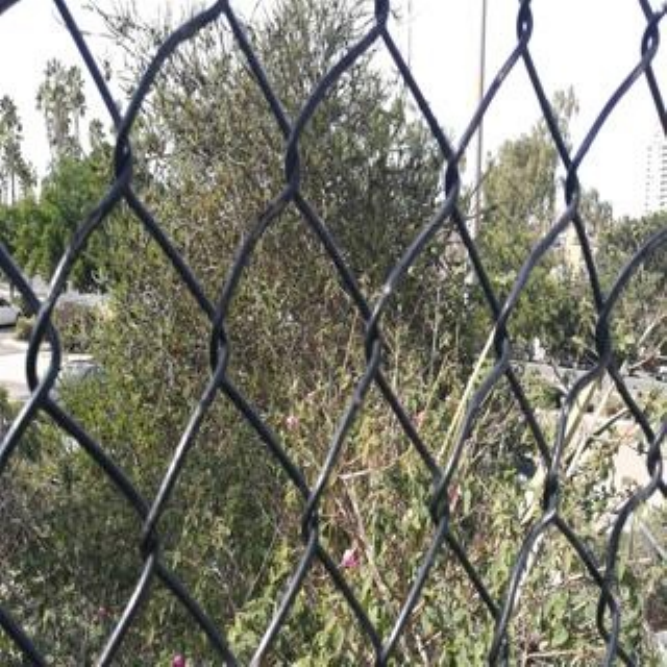}&\hspace{-12pt}
			\includegraphics[width=2.0cm]{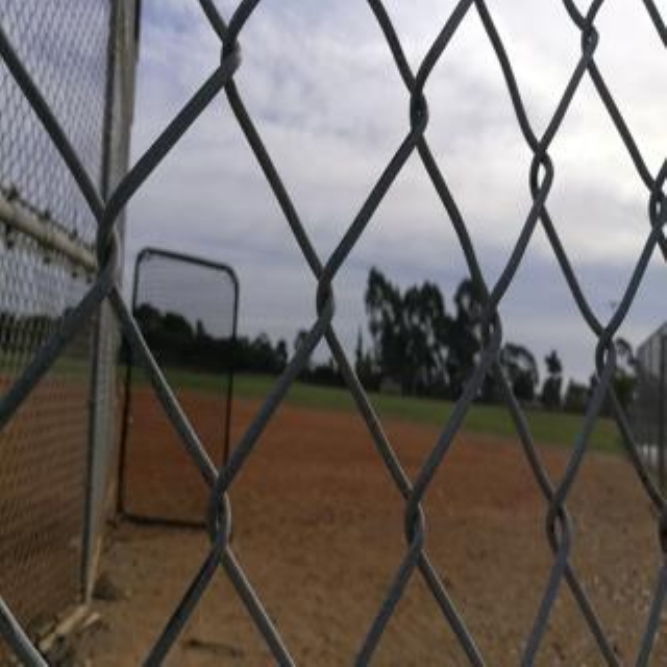}&\hspace{-12pt}
			\includegraphics[width=2.0cm]{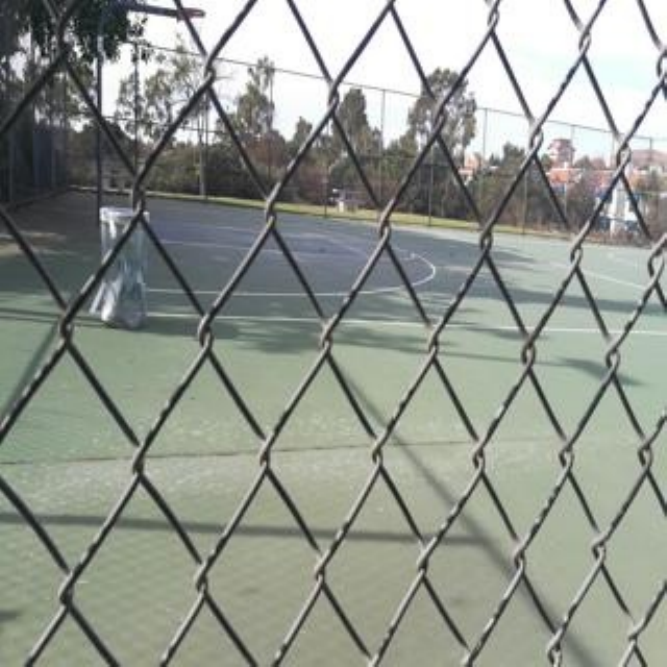}&\hspace{-12pt}
			\includegraphics[width=2.0cm]{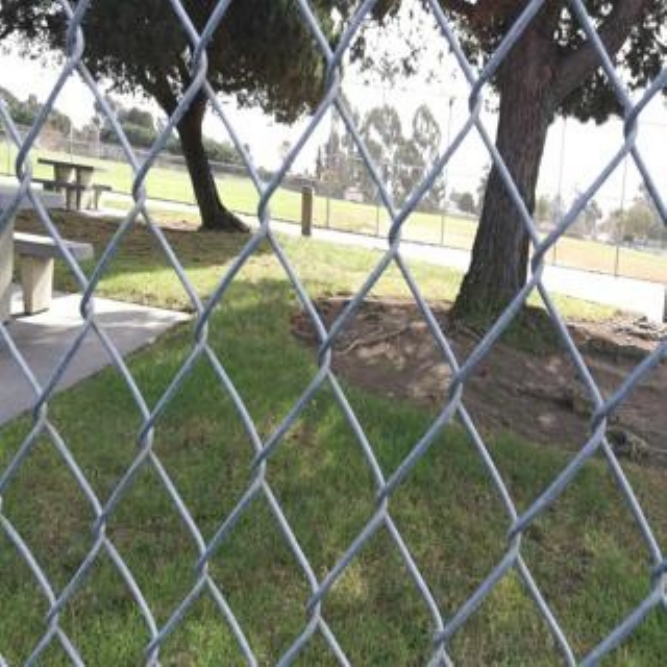}&\hspace{-12pt}
			\includegraphics[width=2.0cm]{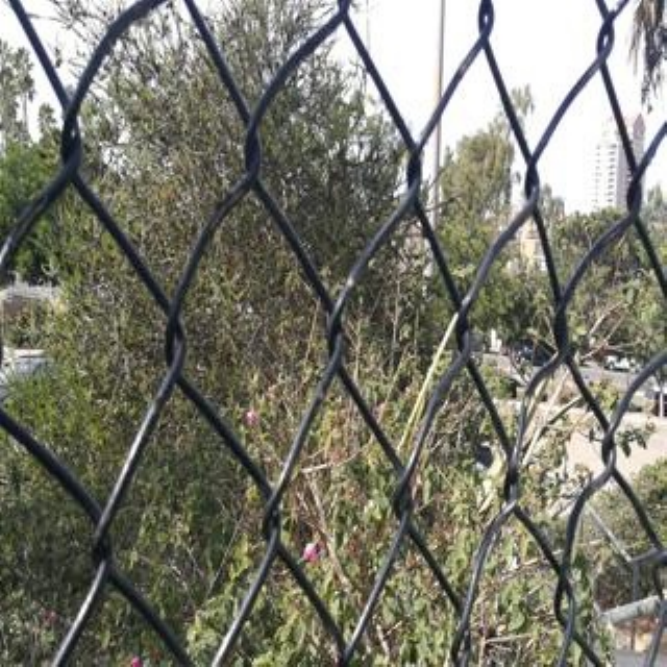}&\hspace{-12pt}
			\includegraphics[width=2.0cm]{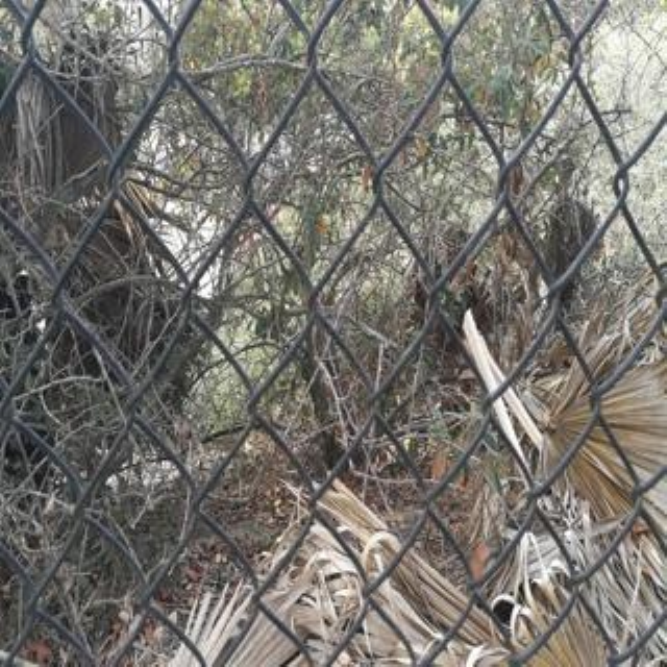}&\hspace{-12pt}
			\includegraphics[width=2.0cm]{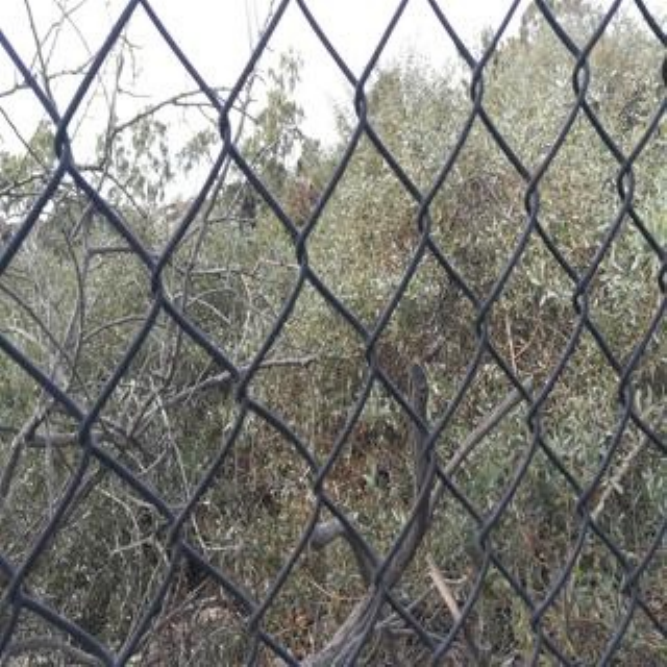}&\hspace{-12pt}
			\includegraphics[width=2.0cm]{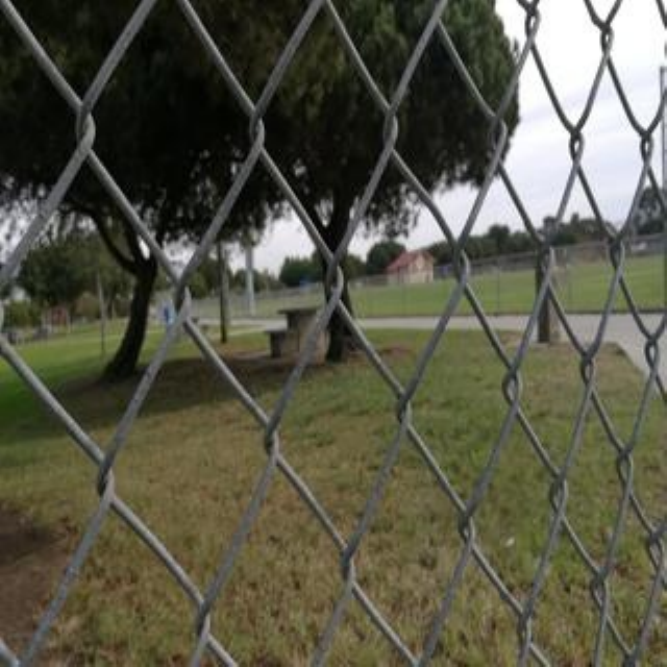}\\
			
			\includegraphics[width=2.0cm]{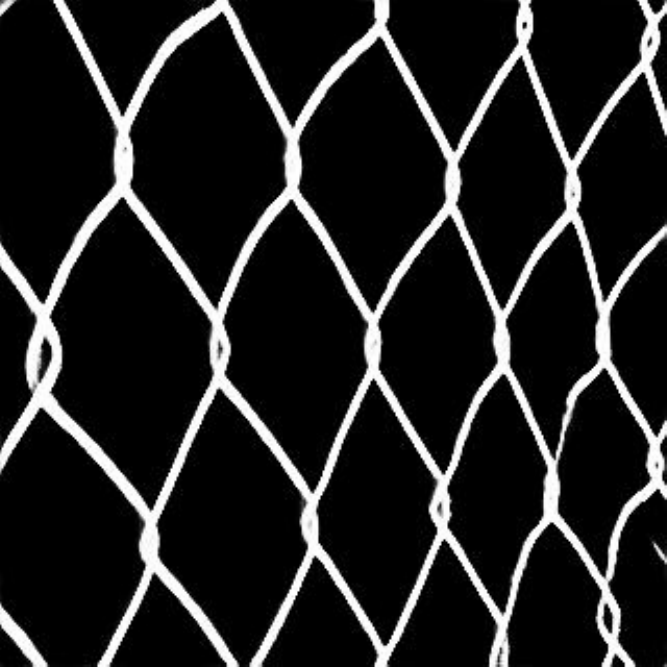}&\hspace{-12pt}
			\includegraphics[width=2.0cm]{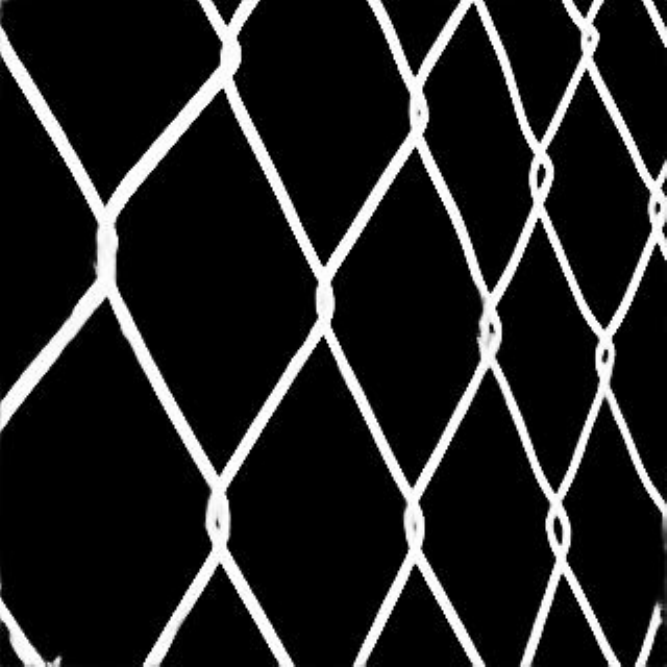}&\hspace{-12pt}
			\includegraphics[width=2.0cm]{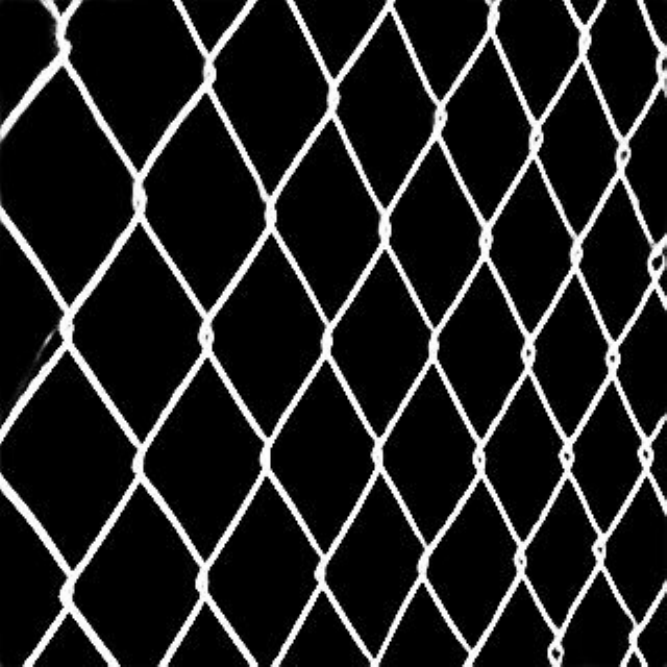}&\hspace{-12pt}
			\includegraphics[width=2.0cm]{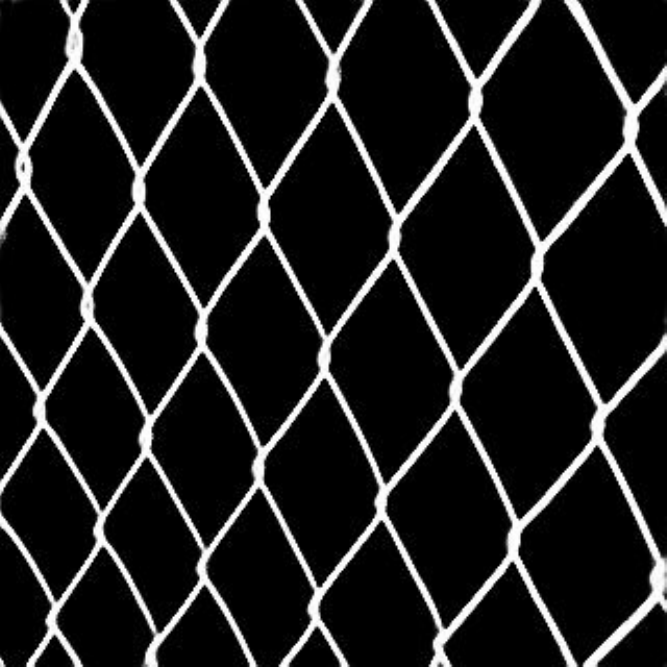}&\hspace{-12pt}
			\includegraphics[width=2.0cm]{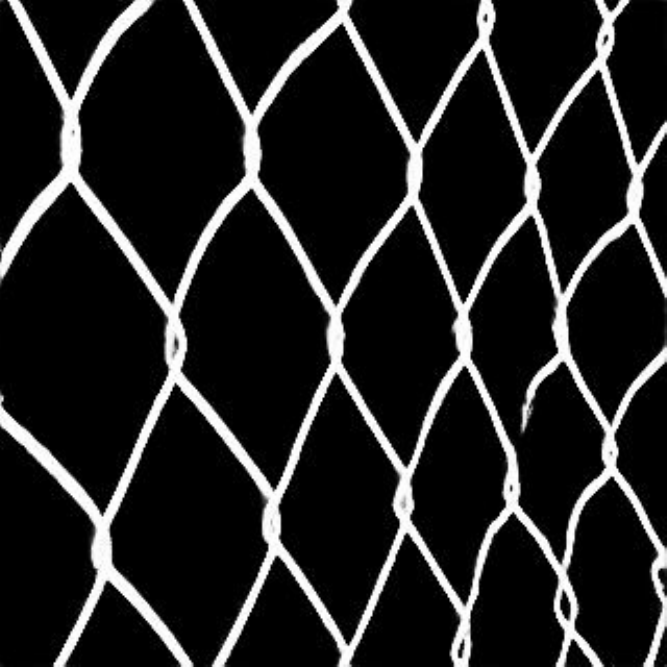}&\hspace{-12pt}
			\includegraphics[width=2.0cm]{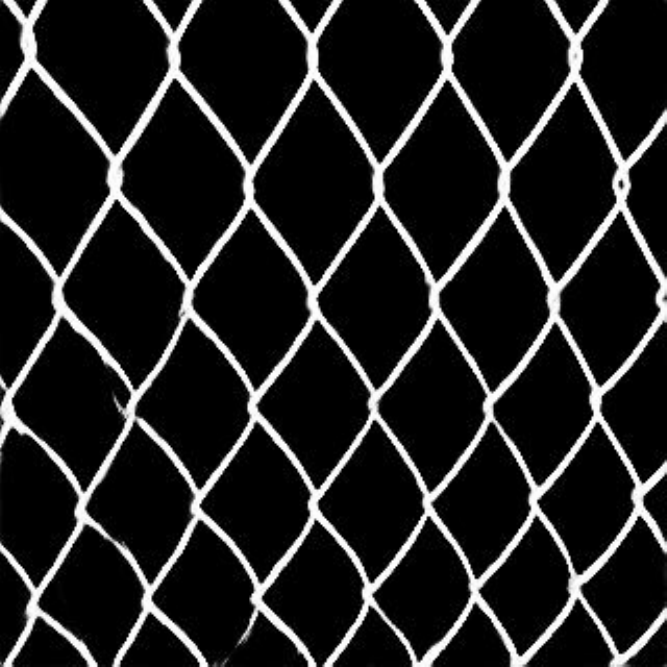}&\hspace{-12pt}
			\includegraphics[width=2.0cm]{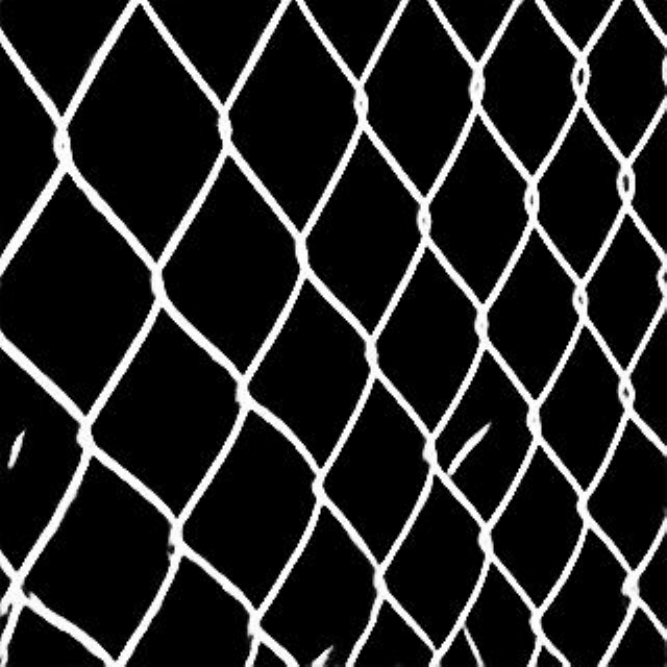}&\hspace{-12pt}
			\includegraphics[width=2.0cm]{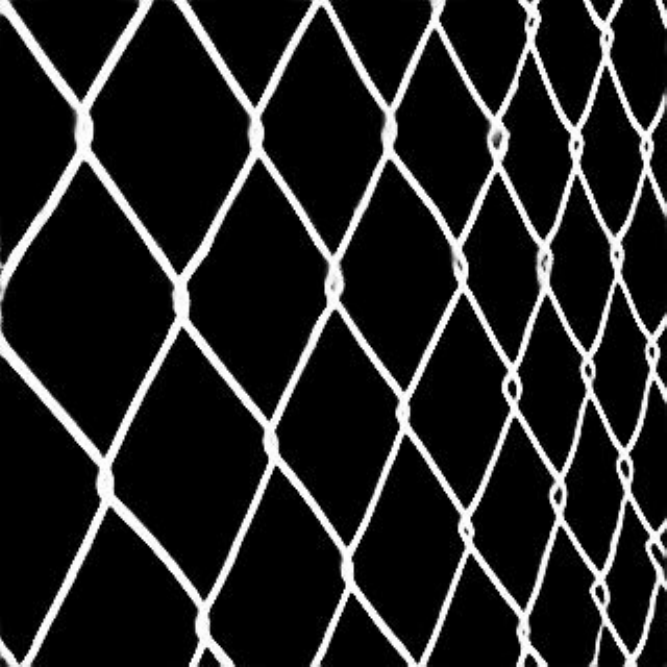}\\
			
			\includegraphics[width=2.0cm]{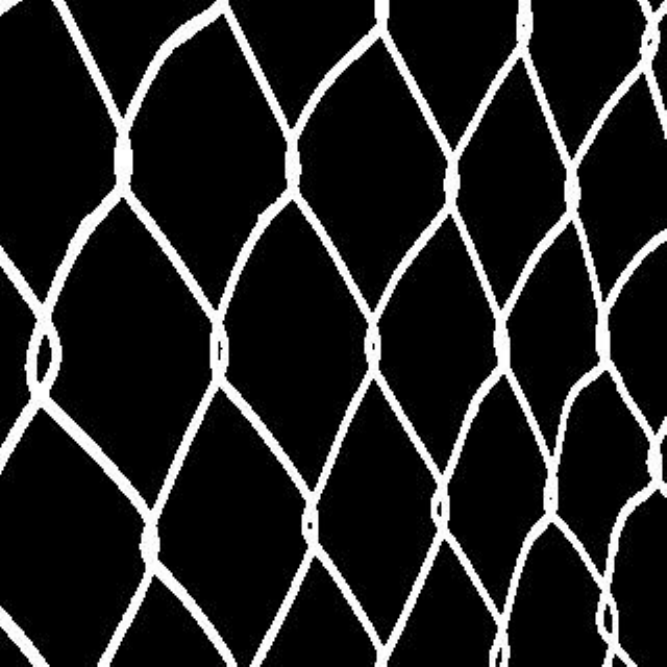}&\hspace{-12pt}					
			\includegraphics[width=2.0cm]{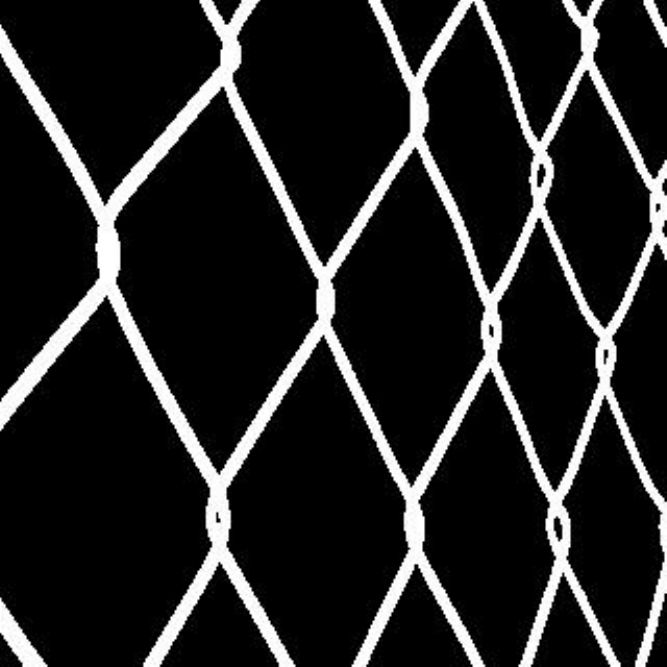}&\hspace{-12pt}					
			\includegraphics[width=2.0cm]{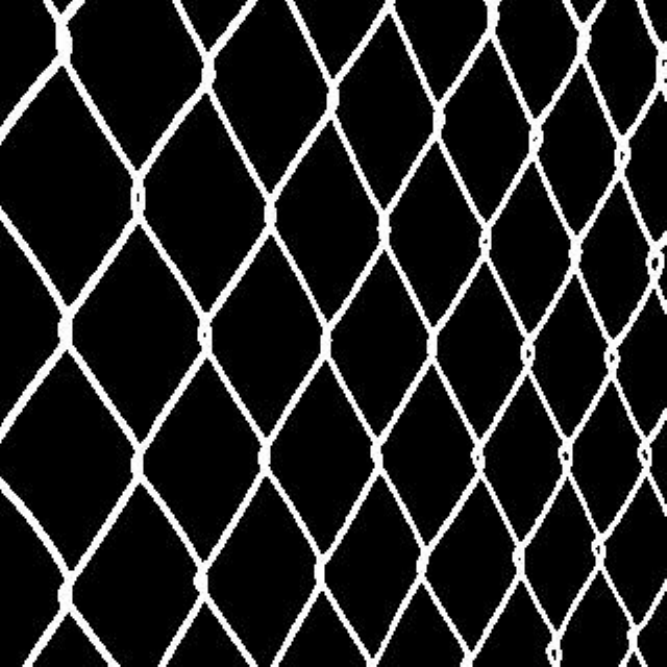}&\hspace{-12pt}					
			\includegraphics[width=2.0cm]{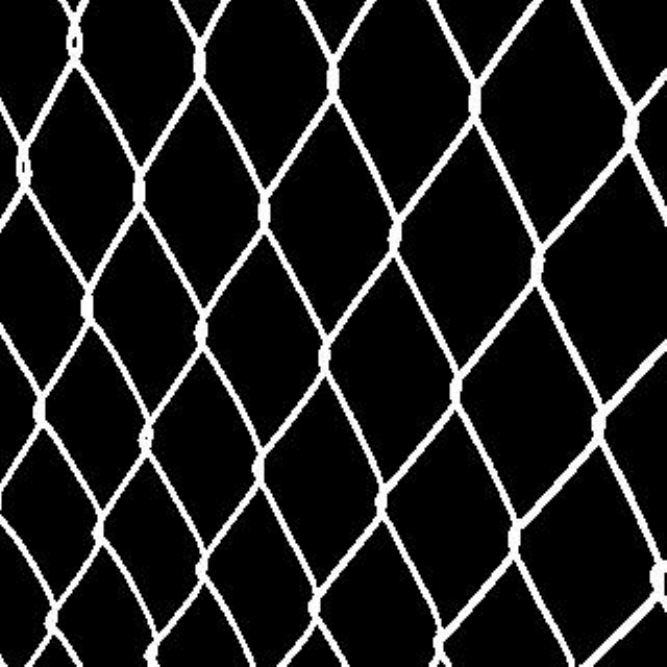}&\hspace{-12pt}					
			\includegraphics[width=2.0cm]{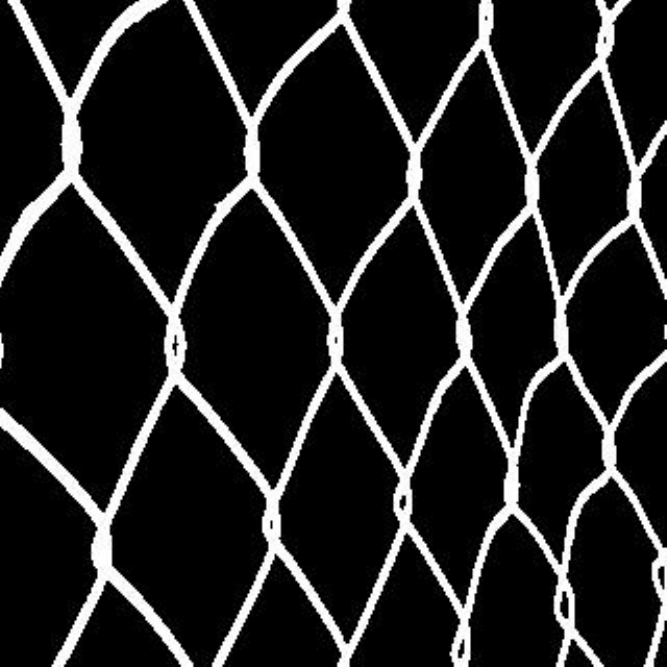}&\hspace{-12pt}					
			\includegraphics[width=2.0cm]{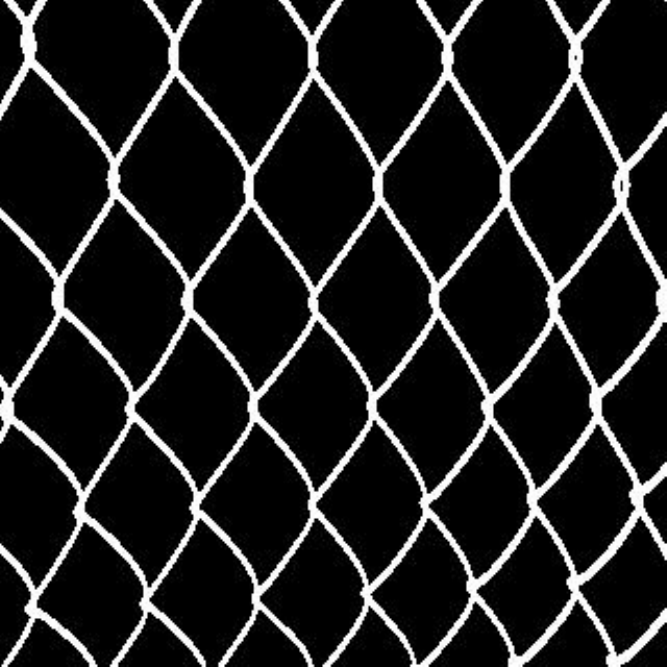}&\hspace{-12pt}					
			\includegraphics[width=2.0cm]{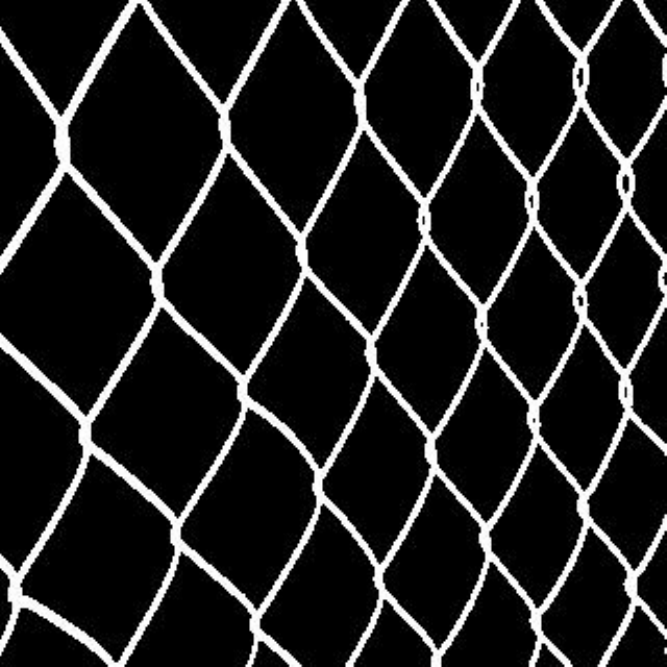}&\hspace{-12pt}					
			\includegraphics[width=2.0cm]{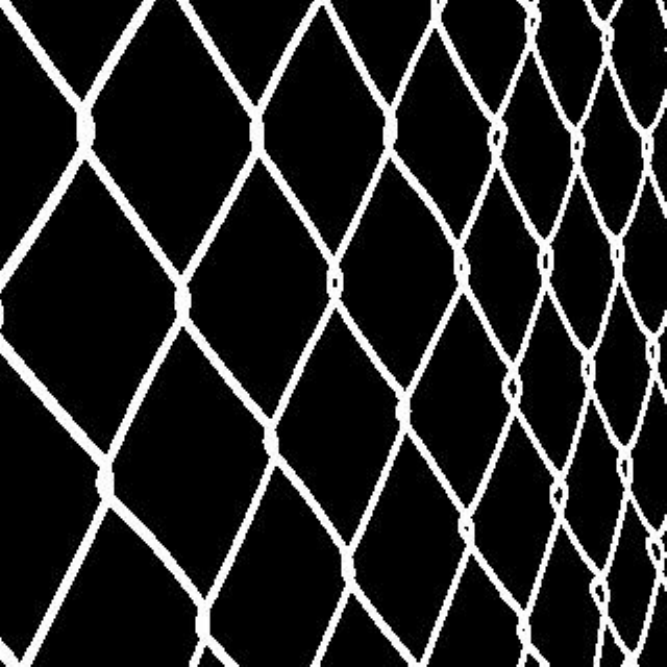}\\					
		\end{tabular}}
		\caption{Fence detection results on UCSD test dataset \protect \cite{Du_2018_ICME}. First row: Input occluded observations. Second row: Fence segmentation results obtained using the proposed model. Third row: Ground truth binary segmentation masks.}	
		\label{fig:fenceSeg}
	\end{figure*}

We show the visual results obtained using the proposed model on the UCSD\_Fence dataset \cite{Du_2018_ICME} in Fig.\ref{fig:fenceSeg}. In the first row of Fig. \ref{fig:fenceSeg}, we show observations degraded by fence occlusions. The results obtained from the proposed framework and the corresponding ground truth fence masks are shown in the second and third rows, respectively. In most   cases, the fence detection results predicted by  the proposed model closely match the ground truth, except for a few samples. For example, the fence mask generated using OccNet for the fenced image in the seventh column of Fig. \ref{fig:fenceSeg} contains some erroneous detections due to fence-like structures in the background.

\begin{table}[!htb]
	\centering
	\caption{Quantitative comparison of fence segmentation algorithms on the proposed IITKGP\_Fence test dataset. The best results are shown in blue color.}
	\scalebox{1.1}
	{
		\begin{tabular}{c c c c c}
			\hline
			\multirow{2}{*}{Method} & \multicolumn{4}{c}{IITKGP\_Fence Dataset} \\\cline{2-5}
			& Precision & Recall & F-measure & MAE \\
			\hline
			FCN \cite{Shelhamer_2017_PAMI} & 0.9529 & 0.9031 & 0.9409 & 0.0012 \\
			UNet \cite{Ronneberger_2015_MICCAI} & 0.9539 & 0.8980 & 0.9404 & 0.0011 \\
			UNet\_Res & 0.9550 & 0.9085 & 0.9439  & 0.0038 \\
			UNet+Adv & 0.9550 & 0.9085 & 0.9439  & 0.0038 \\			
			Proposed & {\color{blue}\textbf{0.955}} & {\color{blue}\textbf{0.9019}} & {\color{blue}\textbf{0.9520}}  & {\color{blue}\textbf{0.0031}} \\			
			\hline
	\end{tabular}}
	\label{tab:quant_fence1}
\end{table}

\begin{table}[!htb]
	\centering
	\caption{Quantitative comparison of fence segmentation algorithms on UCSD\_Fence test dataset \protect \cite{Du_2018_ICME}. The best results are shown in blue color.}
	\scalebox{1.1}
	{
		\begin{tabular}{c c c c c}
			\hline
			\multirow{2}{*}{Method} & \multicolumn{4}{c}{UCSD\_Fence Dataset} \\\cline{2-5}
			& Precision & Recall & F-measure & MAE \\
			\hline
			FCN \cite{Shelhamer_pami2017}  & 0.910 & 0.959 & 0.934 & --\\
			UNet \cite{Ronneberger_2015_MICCAI} & 0.953 & 0.898 & 0.940 & 0.001 \\
			UNet\_Res & 0.955 & 0.908 & 0.943  & 0.003 \\
			UNet + Adv & 0.955 & 0.908 & 0.943  & 0.003 \\
			Proposed & {\color{blue}\textbf{0.955}} & {\color{blue}\textbf{0.917}} & {\color{blue}\textbf{0.946}}  & {\color{blue}\textbf{0.003}} \\			
			\hline
	\end{tabular}}
	\label{tab:quant_fence2}
\end{table}


\subsection{Image inpainting}

\subsubsection{Datasets} We trained and tested our proposed inpainting model using sing a combination of the Places2 \cite{Zhou_2018_PAMI} and CelebA faces \cite{Liu_2015_ICCV} datasets.

\noindent \\
\textbf{Places2 dataset \cite{Zhou_2018_PAMI}:} We trained our network on the training set of Places2 dataset \cite{Zhou_2018_PAMI}. 
The dataset consists of natural images of a variety of scenes and was originally meant for scene classification. The dataset for the fences included $150$ binary images of a variety of fences. The images were resized to a dimension of $256\times 256$ during training. The masked occlusion regions were  replaced by the value of the average pixel in the entire training dataset, which is a constant value for all the input images. Training was done using GPU with a batch-size of $4$. 
 
\noindent \\
\textbf{CelebA faces dataset \cite{Liu_2015_ICCV}: } We also train the proposed inpainting model using the CelebA faces dataset \cite{Liu_2015_ICCV}. It consists of $2,02,599$ images. We follow the splitting scheme provided by the authors \cite{Liu_2015_ICCV}. We used around $19,000$ images for testing, and the remaining images were used  for training the completion network. In order to show the generalization ability of the proposed model  on the faces dataset, we trained using free-form masks, which were taken from \cite{Liu_2018_ECCV}.

\begin{figure*}[!hbt]
	\centering
	\scalebox{0.9}{
		\begin{tabular}{c c c c c c c}
			\includegraphics[width=16.0cm]{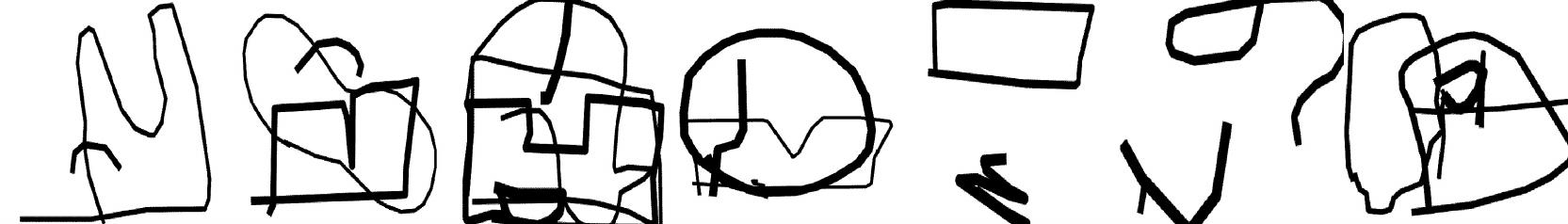}\\			
	\end{tabular}}
	\caption{Examples of free-form masks from \protect \cite{Liu_2018_ECCV}.}
	\label{fig:free-form-masks}	
\end{figure*}

\begin{figure*}[!ht]
	\centering
	\scalebox{0.90}{
		\begin{tabular}{c c c c c c c}
			\includegraphics[width=18cm,height=12cm]{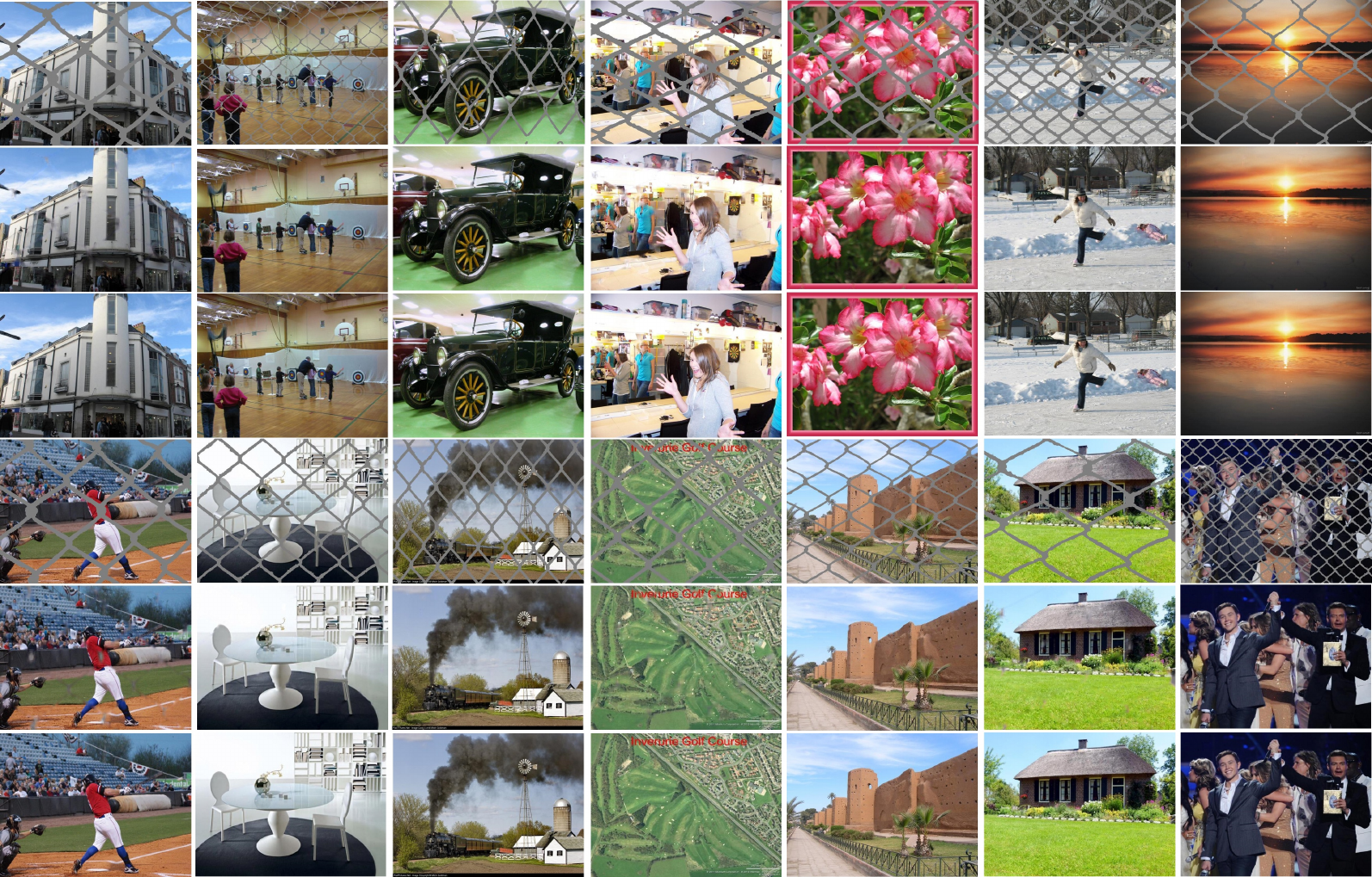}\\					
	\end{tabular}}
	\caption{Image inpainting results on the Places2 dataset \cite{Zhou_2018_PAMI}. Rows 1st and 4th: Input occluded observations. De-fenced results are shown in rows 2 and 4. Rows 3 and 6: Ground truth images.}
	\label{fig:res_def}	
\end{figure*}

\subsubsection{Training Implementations}

\begin{figure*}[!hbt]
	\centering
	\scalebox{1}{
		\begin{tabular}{c}
			\includegraphics[width=14.2cm]{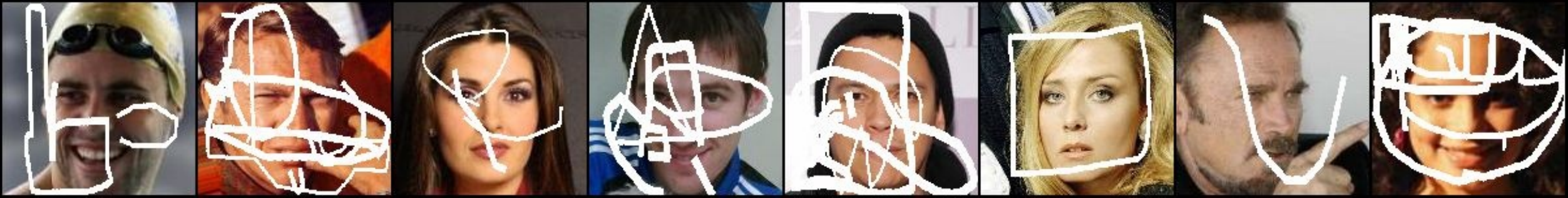}\\
			\includegraphics[width=14.2cm]{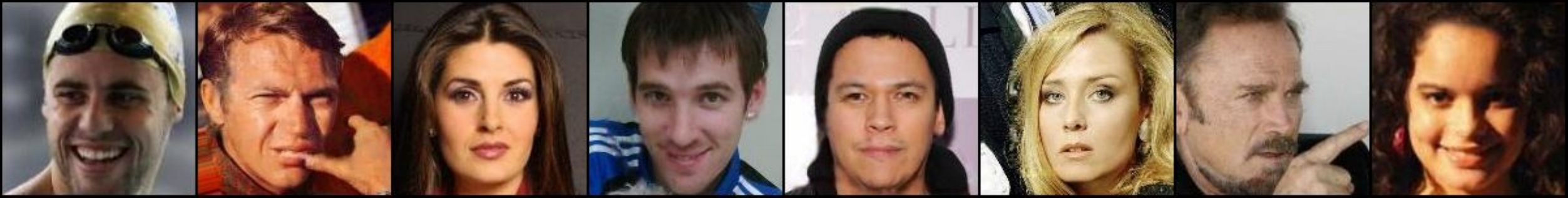}\\
			\includegraphics[width=14.2cm]{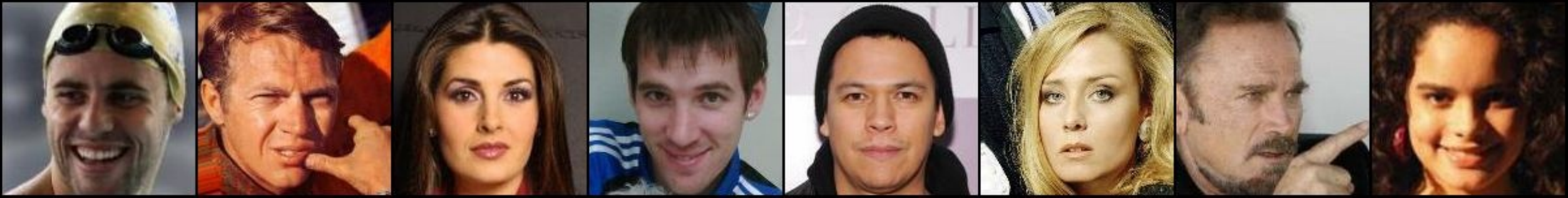}\\						
			\includegraphics[width=14.2cm]{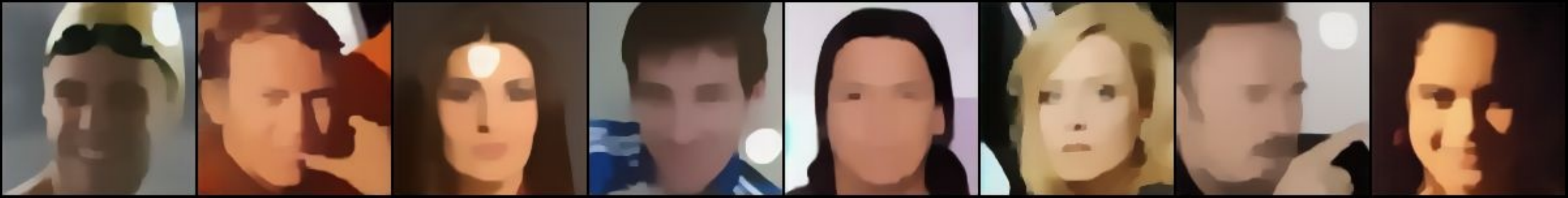}\\
			\includegraphics[width=14.2cm]{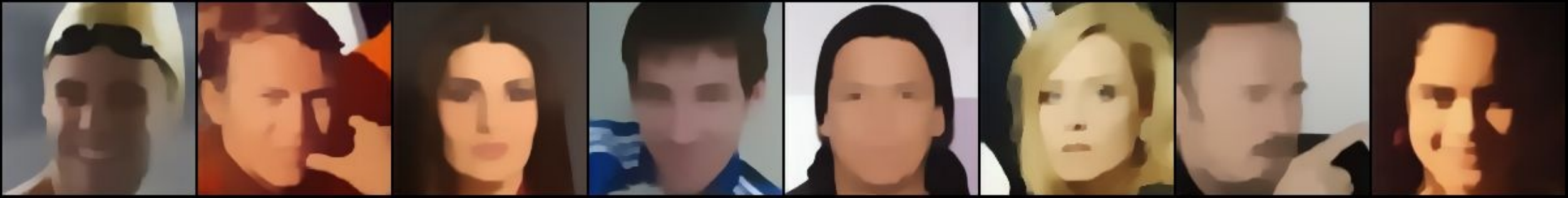}\\
			\includegraphics[width=14.2cm]{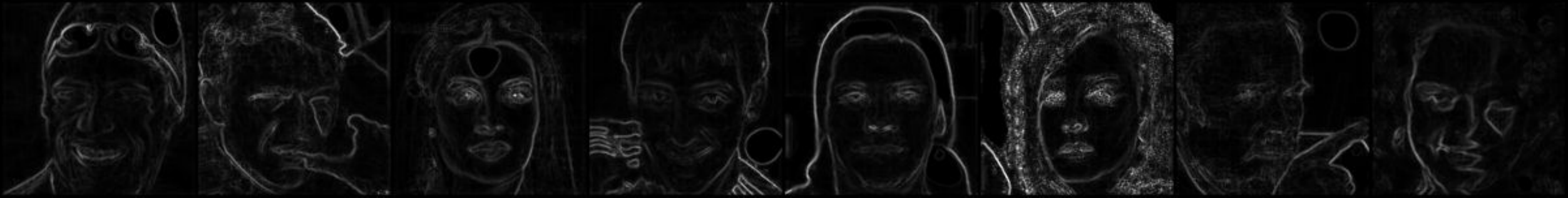}\\
			\includegraphics[width=14.2cm]{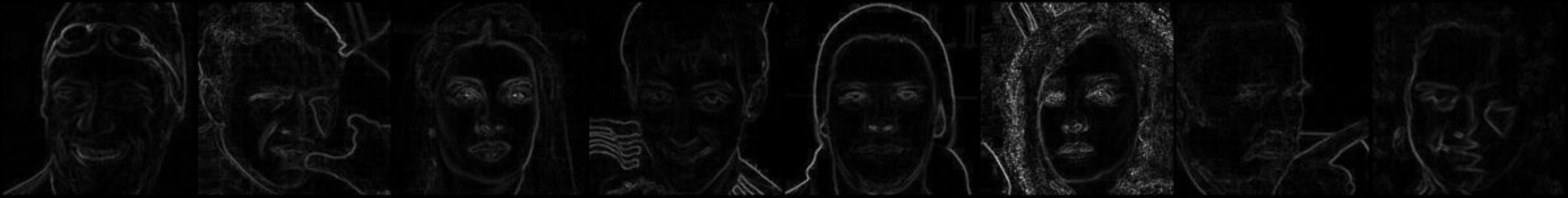}\\
	\end{tabular}}
	\caption{Image inpainting results on CelebA dataset \protect \cite{Liu_2015_ICCV}.  The first row shows the input observations with free-form occlusions. The inpainted and ground truth images are shown in rows $2$ and $3$. The predicted and ground truth image structures obtained using the pre-trained model in \protect \cite{Fan_2019_PAMI} are shown in rows 4 and 5. Finally, rows six and seven show the edge maps from the predicted and ground truth images.}
	\label{fig:celebA_2}	
\end{figure*}

We set the learning rate to $1 \times 10^{-4}$. We optimized the minimax game using the Adam optimizer and trained for 50 epochs. The weighing parameters $\lambda_1$, $\lambda_2$, $\lambda_3$, $\lambda_4$, and $\lambda_5$ in Eq. \ref{eq:loss} were set to $1$, $0.01$, $1$, $0.1$, and $0.1$, respectively.

\subsubsection{Evaluation:} We evaluated the proposed inpainting model on the test split from the Places2 dataset \cite{Zhou_2018_PAMI}. The images with fence-like occlusions are shown in the first and fourth rows of Fig. \ref{fig:res_def}. In the second and fourth rows of Fig. \ref{fig:res_def}, we present image de-fencing results obtained using our trained model. Ground truth images without fence occlusions are depicted in the third and sixth rows of Fig. \ref{fig:res_def}. Note that proposed method produces de-fencing results without any artifacts.

\begin{figure*}[!hbt]
	\centering
	\scalebox{1.0}{
		\begin{tabular}{c}
			\includegraphics[width=14cm]{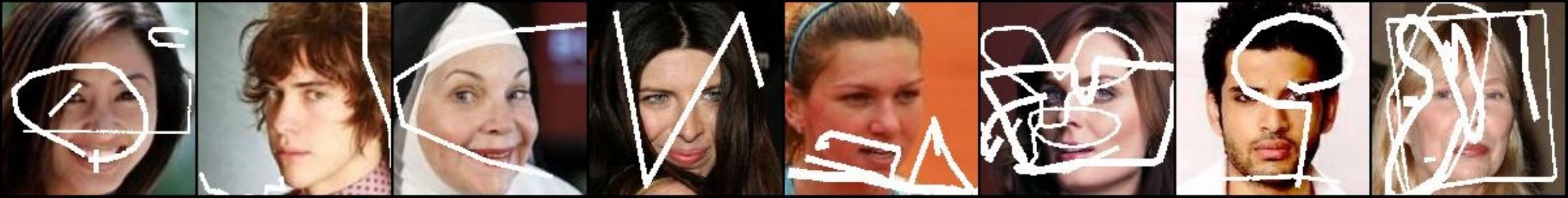}\\
			\includegraphics[width=14cm]{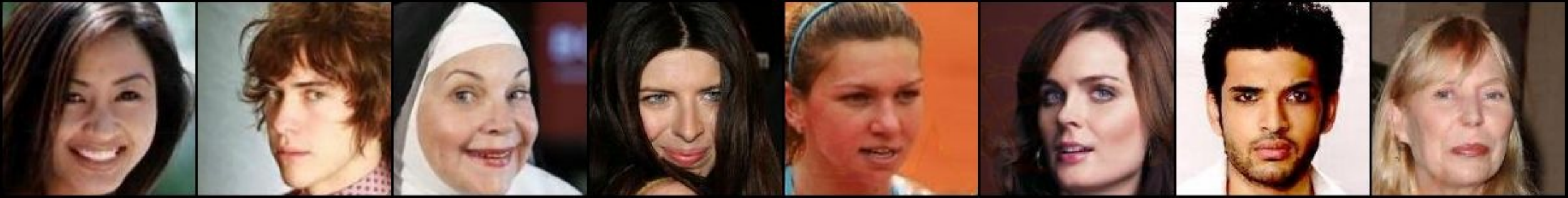}\\
			\includegraphics[width=14cm]{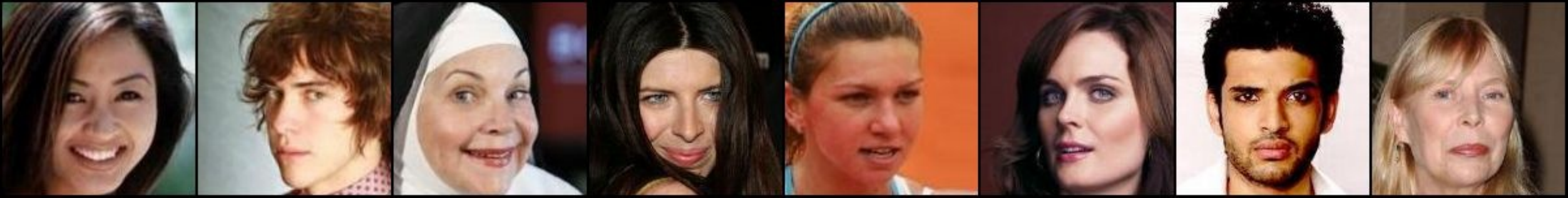}\\						
			\includegraphics[width=14cm]{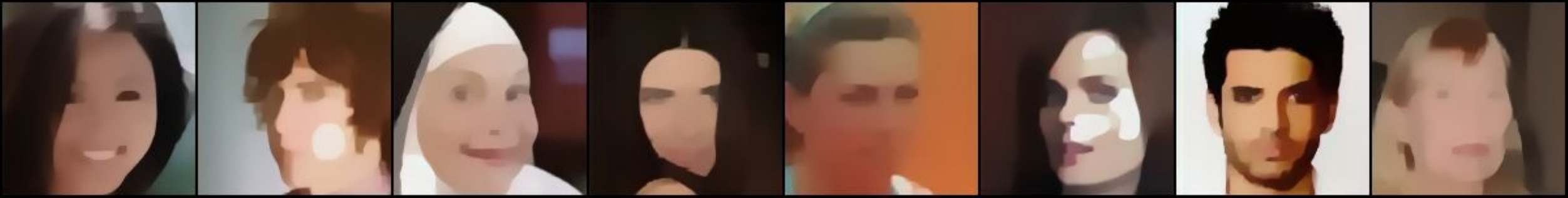}\\
			\includegraphics[width=14cm]{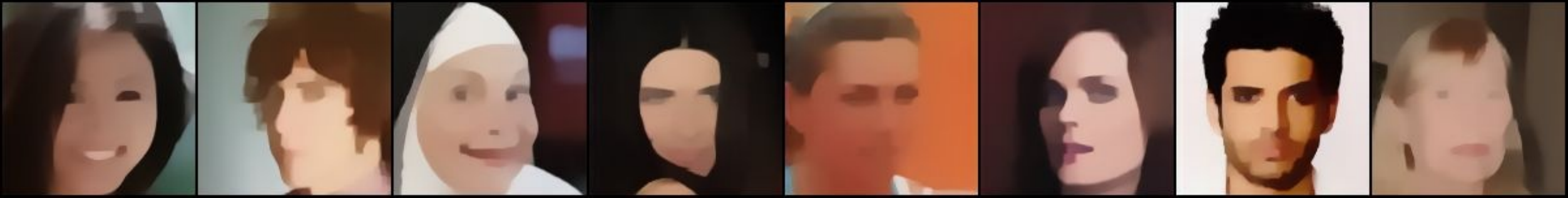}\\
			\includegraphics[width=14cm]{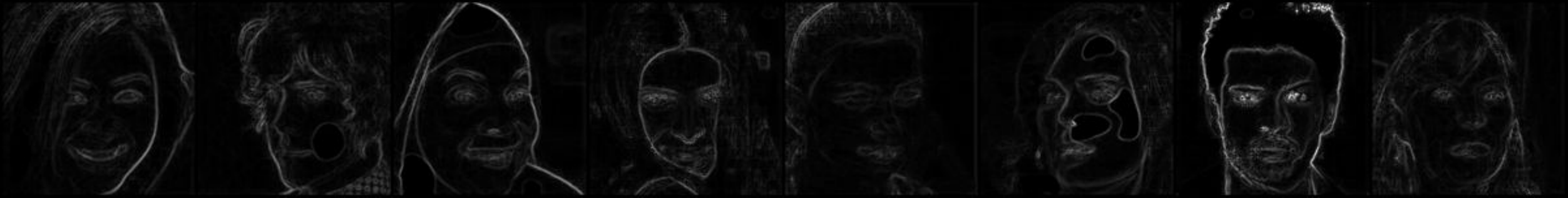}\\
			\includegraphics[width=14cm]{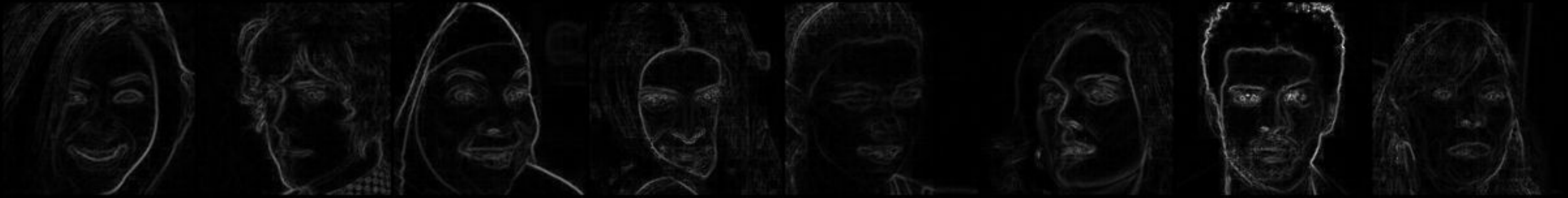}\\
	\end{tabular}}
	\caption{Image inpainting results on CelebA dataset \protect \cite{Liu_2015_ICCV}. The first row shows input observations with free-form occlusions. The inpainted and ground truth images are shown in rows $2$ and $3$. The predicted and ground truth image structures obtained using the pre-trained model in \protect \cite{Fan_2019_PAMI} are shown in rows 4 and 5. Finally, rows six and seven show the edge maps from the predicted and ground truth images.}
	\label{fig:celebA_3}	
\end{figure*}

We also evaluated on the test split from CelebA dataset \cite{Liu_2015_ICCV} to analyze the inpainting performance on generatalized, irrelgular occlusions. The visual results are shown in Figs. \ref{fig:celebA_2} and \ref{fig:celebA_3}. The images with free-form occlusions are shown in the first rows of Figs. \ref{fig:celebA_2} and \ref{fig:celebA_3}. The inpainted images obtained using the proposed inpainting model and the corresponding ground truth images are shown in the second and third rows, respectively. The structures corresponding to the inpainted and the  ground truth images are shown in rows $4$ and $5$, respectively. 
Our trained model generates artifact-free inpainted results, demonstrating its potential and effectiveness on generalized occlusions of any shape.

\section{Conclusion}

In this paper, we proposed a two-stage system for occlusion removal from a single image. Initially, we introduce an encoder-decoder like network to separate occlusions from a given occluded image. Subsequently, we designed a novel image completion generative adversarial architecture that simultaneously recovers structures and textures.  To demonstrate its effectiveness, we evaluate the  model on both fence-like and free-form occlusions. The proposed inpainting model takes input images with free-form masks as inputs and generates inpainted images as outputs.
During evaluation, the trained model  de-fences or inpaints   occluded images in a single step, without the need for  any iterative postprocessing refinement.

{\small
		\bibliographystyle{IEEEtran}
	\bibliography{Ref1,Ref2,Ref3}
}

\end{document}